\documentclass[10pt,journal,compsoc]{IEEEtran}
\usepackage{amssymb,amsmath}
\usepackage{amsfonts}
\usepackage[pdftex]{graphicx}
\usepackage[usenames,dvipsnames]{color}
\usepackage{url}
\usepackage{float}
\usepackage[justification=centering]{subfig}
\usepackage{epstopdf}
\usepackage{cancel}

\ifCLASSOPTIONcompsoc
  \usepackage[nocompress]{cite}
\else
  \usepackage{cite}
\fi
\ifCLASSINFOpdf
\else
\fi
\begin{document}

\title{Effective Backscatter Approximation for Photometry in Murky Water}

\author{Chourmouzios Tsiotsios,
        Maria E. Angelopoulou,
        Andrew J. Davison,
       Tae-Kyun Kim
\IEEEcompsocitemizethanks{\IEEEcompsocthanksitem C. Tsiotsios and A.J. Davison are with the Department of Computing, Imperial College London, SW72AZ, UK.
\IEEEcompsocthanksitem T.K. Kim and M.E. Angelopoulou are with the Department
of Electrical and Electronic Engineering, Imperial College London,
SW72AZ, UK. 
\IEEEcompsocthanksitem E-mails: \{c.tsiotsios,m.angelopoulou,a.davison,tk.kim\}@imperial.ac.uk}
\thanks{Manuscript received ; revised .}}

%
%

\markboth{Journal of \LaTeX\ Class Files,~Vol.~13, No.~9, September~2014}%
{Shell \MakeLowercase{\textit{et al.}}: Bare Demo of IEEEtran.cls for Computer Society Journals}
%



\IEEEtitleabstractindextext{%
\begin{abstract}
Shading-based approaches like Photometric Stereo assume that the image formation model can be effectively optimized for the scene normals. However, in murky water this is a very challenging problem. The light from artificial sources is not only reflected by the scene but it is also scattered by the medium particles, yielding the \emph{backscatter} component. Backscatter corresponds to a complex term with several unknown variables, and makes the problem of normal estimation hard. In this work, we show that instead of trying to optimize the complex backscatter model or use previous unrealistic simplifications, we can approximate the per-pixel backscatter signal directly from the captured images. Our method is based on the observation that backscatter is saturated beyond a certain distance, i.e. it becomes scene-depth independent, and finally corresponds to a smoothly varying signal which depends strongly on the light position with respect to each pixel. Our backscatter approximation method facilitates imaging and scene reconstruction in murky water when the illumination is artificial as in Photometric Stereo. Specifically, we show that it allows accurate scene normal estimation and offers potentials like single image restoration. We evaluate our approach using numerical simulations and real experiments within both the controlled environment of a big water-tank and real murky port-waters. 
\end{abstract}

}

\maketitle

\IEEEdisplaynontitleabstractindextext

%
\IEEEpeerreviewmaketitle

\IEEEraisesectionheading{\section{Introduction}\label{sec:introduction}}
Imaging and scene understanding in murky water has many important applications. In the maritime environments of harbors, shore-lines, rivers and lakes, the human activity is significant. Man-made structures, such as platforms, nuclear reactors, and power cables need to be periodically monitored in order to prevent hazardous situations \cite{Ortiz-Particle}. At the same time, research in marine archaeology and biology has dictated the use of effective optical systems in order to detect, evaluate and analyze the condition of important targets underwater \cite{Eustice-Titanic,Treibitz-Corals}.

However, scene reconstruction in murky water is a very hard task. The camera measures not only the scene-reflected light as in pure air, but also the particle-reflected light -- the so-called backscatter component. As a result, the image contrast is low. Employing traditional Stereo or Structure-from-Motion methods in this case is hard \cite{Sarafraz2}. The image degradation causes strong de-featuring effects and hence some type of photometric post-processing of the images is needed in order to restore contrast and recover features. Additionally, all disparity-based methods fail when the object lacks features and they can only recover a low-frequency estimate of the scene's shape.

Shading-based methods on the other hand, like Photometric Stereo, can be used for estimating dense and detailed shape and albedo even for textureless objects. As robotic platforms carry their own light sources anyway to illuminate the scene, applying photometry in murky water is an appealing option. The effectiveness of photometric methods lies on inverse rendering the image formation model; given the measured brightness and the light source characteristics, the scene albedo and normal vector are estimated that predict the most photo-consistent brightness.

However, in murky water the image formation model is complex. The measured brightness bears not only the information about the scene normals and albedo but also about the particle-reflected backscatter component. The backscatter model is a complex function of the medium's coefficients, the scene depth and the light source characteristics and hence it adds important ambiguity to the Photometric Stereo equations that are now hard to optimize directly for the wanted orientation and albedo. For this reason, previous works have used some simplified versions for the backscatter term. Specifically, they have either neglected the backscatter component entirely \cite{Negahdaripour-PS,Negahdaripour-Shading} or assumed that it is independent of the light source position using an unrealistic setup where both the light sources and the camera were placed outside water at infinite distance from each other \cite{Narasimhan-Structured}. 

In this work, we make significant observations for the profile of the backscatter component on the sensor that lead to its effective approximation. We describe that the position of each pixel with respect to a light source (which is irrelevant in pure air) is crucial in scattering media because it affects strongly the amount of the backscatter component. On the other hand, backscatter exhibits low dependence on scene depth (as opposed to diffuse lighting environments such as fog or haze) and specifically it becomes saturated after a small distance from the camera. Overall, the variation of the backscatter signal on the sensor can be attributed only to the smooth variation of the incident light on the particles in front of the camera. Hence, backscatter is also a smooth function that can be easily approximated via regression using the backscatter value of a few pixels on the sensor. In this way, instead of optimizing the complex backscatter model for all of its unknown variables using the Photometric Stereo equations, we propose both a calibrated and an uncalibrated way for estimating the backscatter component directly from the captured images. Subtracting that from the measured brightness then recovers the useful scene-reflected component which can be optimized effectively for orientation and albedo.

In order to evaluate the effectiveness of our method we perform several experiments through both numerical simulations and real murky water conditions in a big water tank. In simulations, we investigate the dependence of the backscatter function on the light source characteristics and the scene depth, and we compare our backscatter approximation with previous approaches. In our controlled experimental setup, we immerse both the camera and light sources into the scattering medium, separated by a small distance as on a real robotic platform underwater. We evaluate the performance of our method over a wide range of controlled scattering levels by adding milk of gradually increasing quantity and we show that our method outperforms related approaches, compensating effectively for the backscatter effect and yielding Photometric Stereo shape recovery results similar to those in clean water. Finally, we implement a Photometric Stereo system on a Remotely-Operated-Vehicle (ROV) and we show reconstruction results for strong scattering conditions in real port water. Our work gives also further potential for uncalibrated single-image restoration in murky water that we demonstrate with real data.


\section{Related Work}
\label{sBackscatter:Related_Work}

\textbf{Diffuse light \& scattering.}  A large amount of work focused on the image formation model when the illumination is diffuse within fog, haze, and mist \cite{He,Nay2,Tarel}, or sub-sea  \cite{Chiang,Schechner-clear}. In this case, all medium particles between the camera and the scene are illuminated equally. As the scene depth increases, the volume of illuminated particles increases as well, and therefore the total light that gets scattered towards the sensor becomes higher. This implies that backscatter is a depth cue.

In our work, illumination originates from artificial light sources as in any shape-from-shading framework. As we describe, the light propagation model is then significantly different; light is further attenuated due to inverse-square law, the scattering particles receive illumination from specific directions, and the illumination profile of the light sources plays a crucial role for the amount of the backscatter component. These factors lead to a different light propagation model for the backscatter component that we examine in this work.

\textbf{Photometric Stereo approaches.} The complex and under-determined nature of Photometric Stereo in scattering media has lead previous approaches to use simplified backscatter models that are unrealistic in typical sub-sea conditions. Narasimhan et al. \cite{Narasimhan-Structured} formulated Photometric Stereo having both the light sources and the camera outside water and separated by an infinite distance. This leads to a simplified model for the backscatter component which neglects inverse square law and eventually the backscatter variation on the sensor with respect to the light source position. Trying to mimic such a setup by placing the light sources at a far-away distance behind the camera within murky water leads to severe degradation for the image contrast \cite{Jaffe}, as it causes strong backscatter and attenuated illumination on the scene. Thus the model of \cite{Narasimhan-Structured} is mainly applicable for experimental conditions where the light sources and the camera can be placed outside water.

Negahdaripour et al. \cite{Negahdaripour-PS} on the other hand neglected the backscatter component entirely under the assumption that it is insignificant with respect to the direct component. This can be achieved when the camera-source baseline is large enough to reduce backscatter significantly, or when the source-scene distance is small enough to create a very strong direct component. However, assuming that backscatter can be neglected entirely is also unrealistic in typical imaging missions underwater. Indeed, it is well known that some systematic options can mitigate the backscatter effect. Having a light source with limited field of view \cite{Treibitz-Fusion}, or increasing the camera-source baseline \cite{Gupta,Jaffe}, decrease the backscatter component as the illuminated medium volume is also decreased. However, such options do not usually remove backscatter entirely. Specifically, the increase in camera-sources baseline or the decrease in the FOV of the sources cannot be applied illimitably because apart from mitigating backscatter it also degrades the incident illumination on the scene and hence it amplifies noise \cite{Gupta}. In every case, the level of backscatter depends on the scattering level of the medium, which is a factor that cannot be manualy controlled.

Our proposed method relaxes the previous limiting assumptions. It doesn't require a large camera-source separation outside water and it takes account of the strong backscatter that is present in typical imaging conditions \cite{Gupta,Treibitz-Active}. It is based on the observation that backscatter becomes saturated, i.e. independent of scene depth after some distance from the camera. Through numerical simulations we show that this is valid for a wide range of imaging conditions (even for small camera-scene distances) and it outperforms previous backscatter approximation methods. 

In a recent work, Murez et al. \cite{Murez} emphasized the impact of forward scattering, apart of backscattering, on Photometric Stereo in murky water. In our work we focus on the backscatter component and not on other effects that can influence photometric methods in murky water, such as forward-scattering. There are various works indicating that backscatter is the dominant degrading effect in murky water. In \cite{Jagger} it was described that the contrast loss in underwater images (caused by backscatter) is dominant compared with the resolution loss caused by forward-scattering. This was demonstrated with objective criteria in \cite{Sch1,Schechner-clear}, and a wide series of real experiments in a water tank in \cite{Mort}. In \cite{Negahdaripour-Shading} a detailed sensitivity analysis examined the importance of the various model parameters in shape-from-shading underwater, concluding that the impact of the forward-scattering parameters is negligible compared with the ones responsible for attenuation and backscatter. Our work examines thoroughly the backscatter component and its relationship with different imaging characteristics for Photometric Stereo setups in murky water. As we also discuss in Sections \ref{sIm:Phot} and \ref{sBackscatter:Conclusions}, our work is complementary with additional modeling such as near-lighting and forward-scattering. 

\textbf{Backscatter estimation methods.} Although several works showed that backscatter has the dominant impact on image quality underwater \cite{Jagger,Mort,Sch1,Schechner-clear}, very few works described how this can be estimated when it originates from artificial light sources and not diffuse environmental light. In \cite{Treibitz-Instant,Treibitz-Active} backscatter mitigation was achieved using polarizers on the light source and the camera, while in \cite{Murez} it was achieved using a barrier filter in front of the camera when the scene fluoresces. The only approach we found that estimates backscatter without external hardware is the method of \cite{Mort3,Mort} which we compare with ours. In these works, it was assumed that backscatter is proportional to a low-pass filtered version of the original image. However, we show that this assumption yields significant errors to Photometric Stereo because it over-estimates the backscatter component. 

\section{Image Formation Model \& Photometric Stereo}
\label{sBackscatter:Background}

\subsection{Attenuation \& Scattering} As light travels within murky water it gets attenuated and scattered. Specifically, a light beam with initial irradiance $I_0$ is attenuated as it travels a distance $d$ according to \cite{Jaffe}:
\begin{equation}
\label{eLight:Boug}
    I_d=I_0\,e^{-cd},
\end{equation}
where $c$ denotes the total attenuation coefficient of the medium which depends on the size and concentration of the scattering particles. If light is emitted from a point-source, inverse-square law is also taken into account: $I_d=I_0\frac{e^{-cd}}{d^2}$.

Scattering is denoted by the angular scattering function $\beta(\phi)$ of the medium. This expresses the ability of a unit volume to scatter light towards a specific direction:  $\beta(\phi) = \frac{I}{I_0}$, where $I$, $I_0$ denote the scattered and incident light radiance, respectively.  The ability of a particle to scatter light towards all directions around it is reflected by the scattering coefficient $b$. The relation between $\beta(\phi)$ and $b$ depends on the properties of the scattering medium \cite{Chandrasekhar}. For example a simplified version used in \cite{Treibitz-Active} under the assumption that particles scatter light isotropically is $\beta(\phi) = \frac{b}{4\pi}$. A more generalized-anisotropic representation used in \cite{Narasimhan-Structured} is:

  \begin{equation}
  \label{eLight:beta_beta}
    \beta(\phi) = \frac{b}{4\pi}(1 + g\cos\phi),
\end{equation}
where $g \in (-1,1)$ is a parameter that depends on the characteristics of the scattering particles. 


 \begin{figure}[tp]%
 	\centering
 	\includegraphics[width=0.33\textwidth]{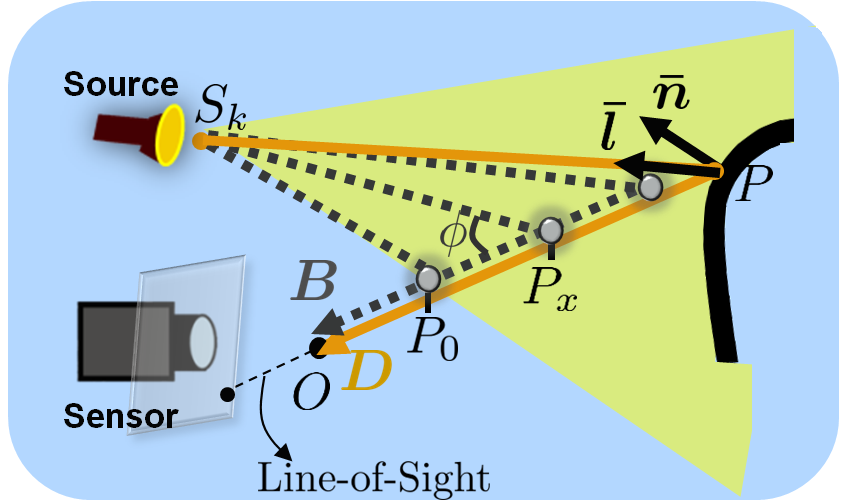}
 	\caption{The brightness at a sensor pixel equals the sum of the attenuated scene-reflected signal (direct component) and the particle-reflected signal (backscatter component).}
 		\label{fBackscatter:image}
 	\end{figure}

\subsection{Image Formation Model} As Figure \ref{fBackscatter:image} shows, the total brightness at a sensor pixel from a source $S_k$ corresponds to the sum of the scene-reflected direct component $D_k$ and the particle-reflected backscatter component $B_k$ (see \cite{Treibitz-Active,Narasimhan-Structured} for further details).

The direct light component is given by the equation:
\begin{equation}
\label{eLight:D}
D_k = I_k\frac{e^{-c(|PS_k| + |OP|)}}{|PS_k|^2} \, \boldsymbol{\hat{l_{PS_k}}}\cdot \boldsymbol{n},
\end{equation}
where $I_k$ denotes the radiant intensity of the source, $|PS_k|$ is the distance between the source and the scene point $P$, and $c$ is the total attenuation coefficient of medium. Here, according to Equation \ref{eLight:Boug}, the term $\frac{e^{-c(|PS_k| + |OP|)}}{|PS_k|^2}$ reflects the attenuation that a light beam suffers as it travels the distance $|PS_k|$ from the source to the scene and then the distance $|OP|$ from the scene to the sensor, due to the medium attenuation and  inverse-square law. According to the shading model, the reflected brightness also depends on the dot product between the incident illumination direction $\boldsymbol{\hat{l_{PS_k}}}$ and the normal vector $\boldsymbol{n}$ of the surface patch. Here, $\boldsymbol{\hat{l_{PS_k}}}$ is a unit vector denoting direction only, and $\boldsymbol{n}$ is a non-unit vector whose magnitude equals the surface patch albedo $|\boldsymbol{n}| \equiv \varrho$. In a more compact representation the direct component can be written as

\begin{equation}
	\label{eLight:Dcompact}
	D_k = \boldsymbol{l_{PS_k}}\cdot \boldsymbol{n}.
\end{equation}
Here the non-unit vector $\boldsymbol{l_{PS_k}}$ bears all the information (direction and magnitude) about the attenuated illumination onto the scene point: $\boldsymbol{l_{PS_k}} \equiv I_k\frac{e^{-c(|PS_k| + |OP|)}}{|PS_k|^2} \, \boldsymbol{\hat{l_{PS_k}}}$.

The backscatter signal corresponds to the light component that travels from the light source to the particles and then it gets scattered towards the image sensor. In order to define its brightness value on the sensor pixel we need to calculate the summation of all light beams along the Line-Of-Sight (LOS) that get scattered towards the pixel. Consider first a differential volume of particles at a point $P_x$ on the LOS of the sensor pixel (Figure \ref{fBackscatter:image}). The differential volume receives an attenuated light signal $I_k\frac{e^{-c|S_kP_x|}}{|S_kP_x|^2}$ from the source (Equation \ref{eLight:Boug}), then scatters only a proportion of this signal towards the sensor according to the angular scattering function $\beta(\phi)$ (Equation \ref{eLight:beta_beta}). The scattered light is also attenuated by $e^{-c|OP_x|}$ as it travels from the particle to the sensor. Thus, the differential amount of backscatter light from a particle volume along the LOS is $dB_k = \frac{b}{4\pi}(1+g\cos\phi) I_k\frac{e^{-c(|S_kP_x|+|OP_x|)}}{|S_kP_x|^2}$. In order to define the total amount of the backscatter component for the sensor pixel, we have to integrate this along all positions on the LOS where particles backscatter light towards the sensor:

\begin{equation}
\label{eLight:B}
B_k = \int_{P_k}^{P} I_k \, 
\frac{b}{4\pi}(1+g \cos \phi)
\, 
\frac{e^{-c(|S_kP_x|+|OP_x|)}}{|S_kP_x|^2}
dP_x.
\end{equation}
Note here that the integration starts at a point $P_k$ along the LOS, which corresponds to the intersection point between the LOS and the limited field of view of the light source $k$. All particles between the origin and $P_k$ are not directly illuminated by the light source and thus they don't backscatter any light towards the pixel. The integration ends at the scene point $P$. There is no closed-form equation for the solution of the integral, but it was shown that it is a smooth function that can be evaluated numerically \cite{Sun}.

\subsection{Photometric Stereo}
\label{sIm:Phot}

The total brightness at every pixel is the sum of the direct and backscatter components from Equations \ref{eLight:Dcompact} and \ref{eLight:B}: 

\begin{equation}
\label{eLight:E}
E_k=D_k+B_k.
\end{equation}
Photometric Stereo aims at recovering the normal vector $\boldsymbol{n}$ (unit normal vector scaled by the albedo $\varrho$) for every pixel, using the measured intensities $E_k$, $k \in [1,noS]$, where $noS$ is the total number of sources. Let us examine the number of unknown variables for each component. 

We assume here that the light positions $S_k$ are fixed and known with respect to the camera as in a typical robotic platform underwater. For the direct component, the unknown variables per-pixel are the normal vector $\boldsymbol{n}$ (scaled by the albedo) and the scene point position $P$. The total attenuation coefficient $c$ is a global unknown for all pixels. As in traditional pure-air Photometric Stereo formulations, the direct component can be simplified assuming that the camera-scene distance is large compared to the object size. Then, lighting can be assumed to be distant, i.e. the incident illumination vector on the scene $\boldsymbol{l_{PS_k}}$ (Equation \ref{eLight:Dcompact}) is constant for all pixels and can be calibrated using a white matte sphere  \cite{Narasimhan-Structured, Tsiotsios}. The distant-lighting assumption can be relaxed as shown in \cite{Kolagani,Negahdaripour-PS,Tsiotsios-CVIU}, assuming that the total attenuation coefficient $c$ and the average camera-scene distance are known. In this case, an iterative algorithm can be applied which estimates an initial version of the scene normals based on the distant-lighting assumption and then gradually refines both the normals and the estimated lighting on the scene, by using the depth map that is recovered when the normal map is integrated. In this work, we do not use such a refinement step as we draw the focus on the dominant backscatter term. The work of \cite{Tsiotsios-CVIU} describes how additional modeling for the direct component can complement our backscatter estimation method.

The backscatter component on the other hand exhibits a more complex form and additional unknowns; apart from the global variables of the medium $c$ and $\beta(\phi)$, it depends on the scene point position $P$ and the position of the minimum lit point $P_k$. Special emphasis should be given to $P_k$, which is an unknown variable of the backscatter component only and adds significant ambiguity to the system of equations. Specifically, this corresponds to the intersection point between each pixel's ray and the beam profile of the light source. Defining $P_k$ per-pixel is hard since apart from the $3D$ position of the source, its rotation and beam angle should be known. Such details are difficult to calibrate accurately \cite{Treibitz-Active}, especially if light sources exhibit non-uniform illumination profile \cite{Park}. Furthermore, in Section \ref{sBackscatter:Simulations} we show that the backscatter function is very sensitive to changes with respect to the light source characteristics, and thus errors in calibration are expected to reflect errors in backscatter estimation and Photometric Stereo reconstruction.  For this reason, previous works have used backscatter model approximations to simplify the problem.

\textbf{Previous backscatter approximations:} In \cite{Kolagani,Negahdaripour-PS,Negahdaripour-Shading} it was assumed that backscatter is negligible with respect to the direct component and thus the photometric equations can be expressed using only the direct component: $E_k = D_k + \cancelto{0}{B_k}$. We refer to this approximation later on as `No Backscatter'. Narasimhan et al. \cite{Narasimhan-Structured} used a simplified model for the backscatter term, using an experimental setup where the light sources were placed outside water at a large distance behind the camera. Using this setup they neglected inverse square law and they assumed that the backscatter component does not depend on the light source position as the illumination from a source on all scattering particles is distant. According to their formulation, when the light source setup is symmetric (equidistant sources from the camera) the backscatter component from every source for a specific pixel should be constant: $B_{k} = B_{k+1}, \, \forall k \in [1,noS-1]$, where $noS$ is the number of sources. Thus, backscatter can be eliminated by subtracting pairs of measured intensities that correspond to different light sources: $E_{k} - E_{k+1} = D_{k} - D_{k+1} + \cancelto{0}{B_{k} - B_{k+1}}, \hspace{4mm} \forall k \in [1,noS-1]$. 

In the next sections we examine more closely the backscatter variation on the sensor with respect to changes in scene depth and light source characteristics. We show that the backscatter saturation with scene distance and its strong dependence on the light source position make it a smoothly varying signal across the sensor that can be easily approximated either via a calibrated or uncalibrated way. 

\section{Backscatter Variation}
\label{sBackscatter:BVariation}

In realistic conditions underwater the camera and light sources are carried by the same body separated by a small offset. The resulting backscatter model (Equation \ref{eLight:B}) depends on many factors that have been neglected in the previous approaches, such as inverse-square light fall-off and also the beam angle and the position of the light sources. In this section, we investigate these factors through numerical simulations and we show that backscatter has a non-uniform profile on the sensor due to the varying pixel position with respect to each source but not due to variation with scene depth. Thus, it is a smooth function on the sensor that differs for every light source.



\subsection{Variation with Light Source and Pixel Position}

The medium particles in front of the camera that cause backscattering toward the sensor are much closer to the light source than the scene. Hence, they receive strongly near-field illumination that changes significantly from one pixel to the other. Previous approximations neglected this physical characteristic.

\begin{figure}%
 	\centering
 	\includegraphics[width=0.45\textwidth]{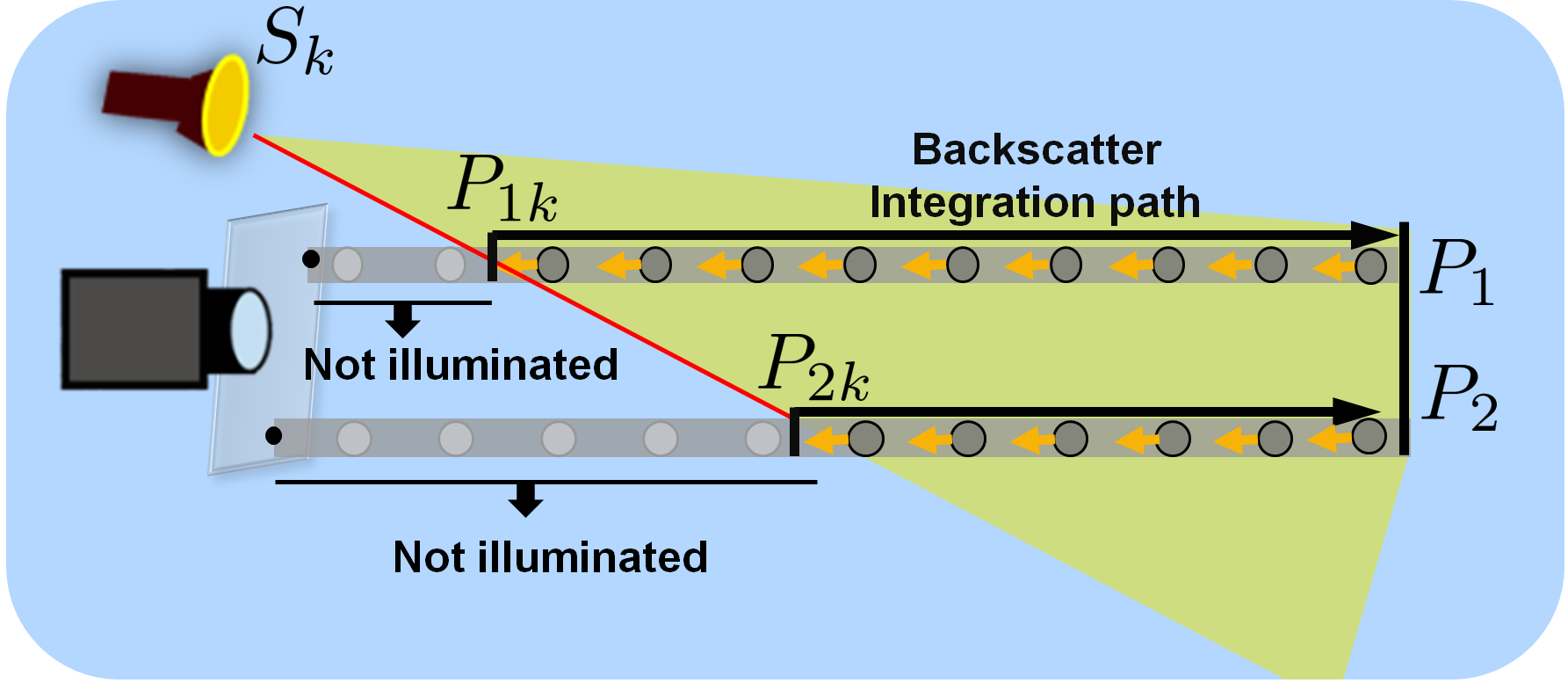}
 	\caption{When a pixel is closer to the light source with respect to another, it has a bigger backscatter integration path as its LOS intersects the FOV of the source at a smaller depth. Furthermore, the particles along its LOS receive and backscatter stronger illumination due to inverse-square light fall-off. Thus, a pixel closer to the light source receives a higher backscatter component.}
 		\label{fBackscatter:B_Variation_1}
 	\end{figure}

Figure \ref{fBackscatter:B_Variation_1} shows the backscatter integration path (Equation \ref{eLight:B}) for two scene points that are illuminated by the same source. First, due to the limited beam angle of the source, the lower limit of the integral will differ for the two pixels. This leads to different integration paths. Specifically, the pixel that is closer to the light source will have a bigger integration path and thus it will receive a larger backscatter component. Second, due to the small distance between the source and the particles, inverse square law cannot be neglected as in \cite{Narasimhan-Structured}. The distance $|S_kP_{1k}|$ will be significantly smaller than $|S_kP_{2k}|$ (and similarly all the distances between the source and the particles along the LOS of each pixel), which will also lead to stronger backscatter component for the pixel closer to the source. Thus, even when two pixels correspond to equidistant scene points, they are expected to receive different backscatter signals.

\begin{figure}%
 	\centering
 	\includegraphics[width=0.4\textwidth]{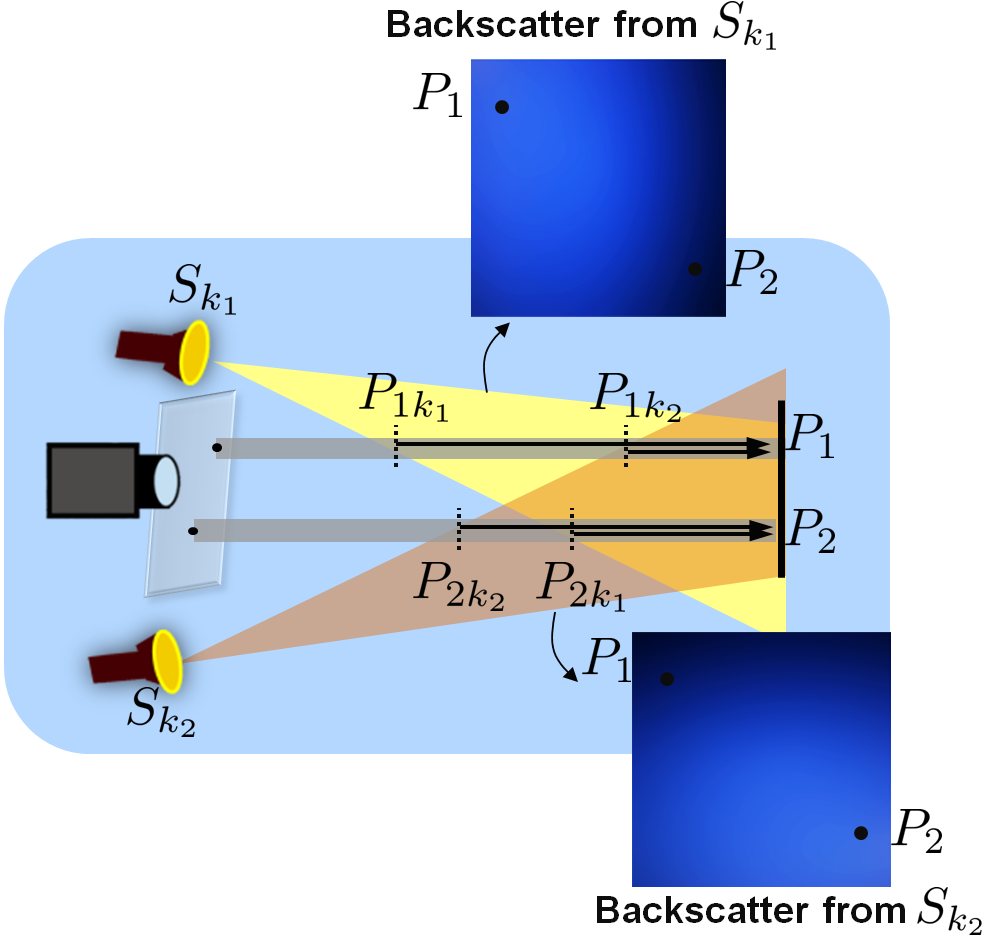}
 	\caption{The backscatter integration path of a sensor pixel differs per source.  For example the pixel that corresponds to the scene point $P_1$ receives significantly stronger backscatter when the light source $k_1$ is used because it is closer to that source. Similarly, $P_2$ receives stronger backscatter when the source $k_2$ is used. Generally,  each source creates a distinct non-uniform backscatter component on the sensor (illustrated by the simulated backscatter images).}
 		\label{fBackscatter:B_Variation_2}
\end{figure}

Naturally,  backscatter doesn't vary only for every pixel but for every source as well (Figure \ref{fBackscatter:B_Variation_2}). When a pixel is closer to a light source $k_i$, its LOS will intersect the beam angle of $k_i$ at a smaller depth point $Z_{k_i}$ than it will intersect the other sources. Furthermore, the particles along the LOS will receive (and scatter) more light due to the smaller distance from the specific source. Every light source finally creates an uneven backscatter component on the sensor according to its position with respect to each pixel. The synthetical backscatter images (using Equation \ref{eLight:B}) of Figure \ref{fBackscatter:B_Variation_2} illustrate this non-uniformity. Contrary to both previous assumptions in \cite{Narasimhan-Structured,Negahdaripour-PS}, the backscatter is strong and it varies per pixel and light source despite that the sources are symmetric with respect to the camera and the scene points are equidistant.

\subsection{Variation with Scene Depth}
As Equation \ref{eLight:B} indicates, backscatter for every pixel is a function of both the position of the minimum lit point $P_k$ and the scene point $P$. The depth values of these two points determine the integration path and the respective intensity of the backscatter per pixel. As described in the previous section, pixels that are closer to a light source will have a bigger integration path and will therefore receive stronger backscattered light. Let us now examine how the scene depth affects the backscatter component.

\begin{figure}%
     \centering
     \subfloat[][Backscatter-Depth ]{\label{fBackscatter:Bz:a}\includegraphics[height=0.095\textheight]{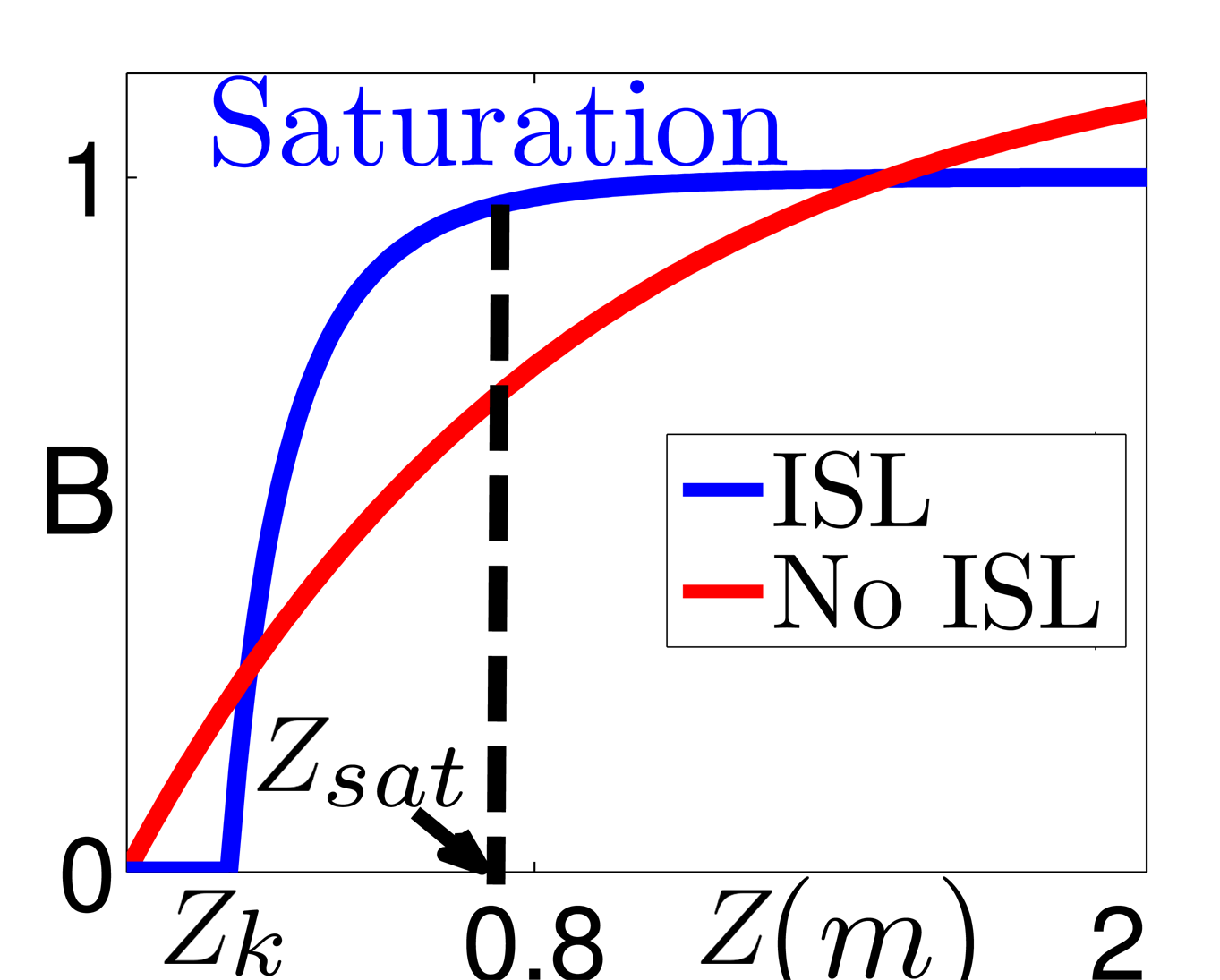}}%
     \hspace{0.05mm}
     \subfloat[][Scene Depth Saturation]{\label{fBackscatter:Bz:b}\includegraphics[height=0.09\textheight]{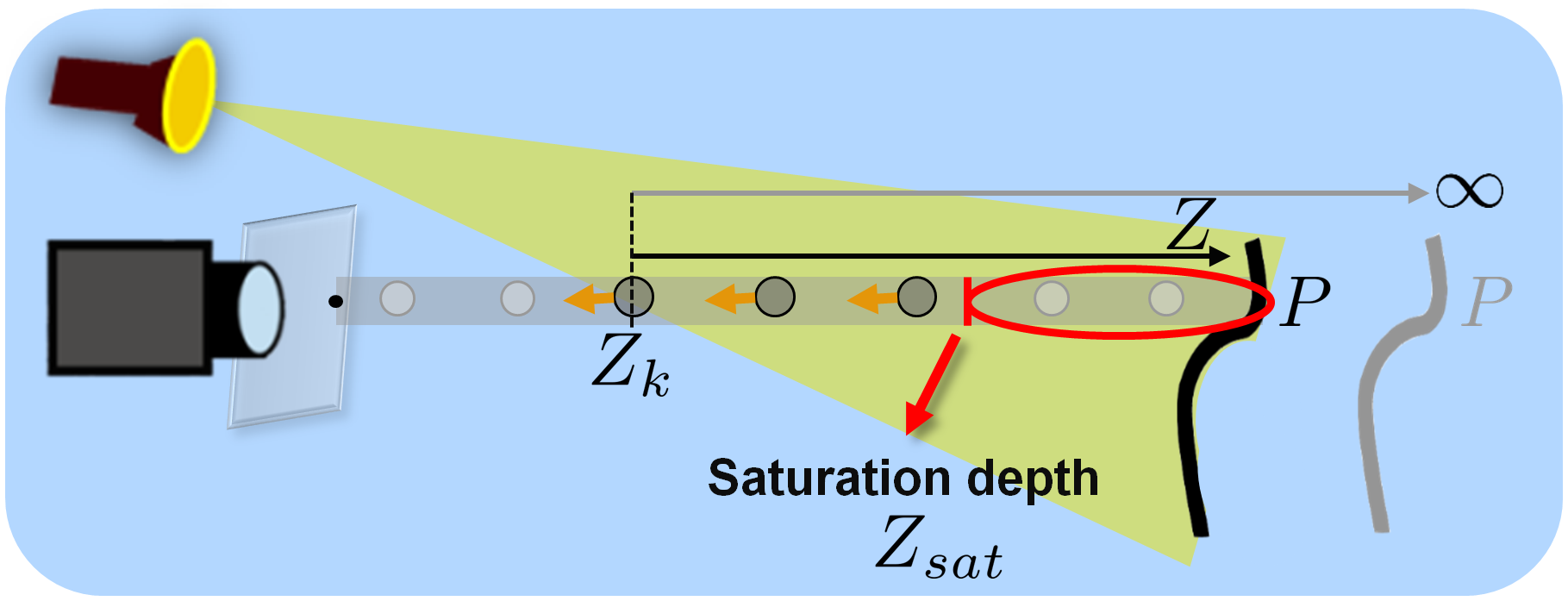}}%
     \vspace{-0mm} 
\caption{Due to inverse-square law, backscatter is saturated after a small depth $Z_{sat}$ away from the camera and hence it captures no information about scene's depth $Z$. }
    \label{fBackscatter:Bz}
\end{figure}

Contrary to cases of diffuse or distant from camera illumination \cite{He,Narasimhan-Structured}, inverse-square law (ISL) is considered when modeling the backscatter component. In this case, the backscatter function is less sensitive to changes in scene depth as it was indicated in \cite{Treibitz-Active}. Figure \ref{fBackscatter:Bz:a} shows the backscatter function for increasing scene depth when ISL is considered. It equals $0$ below the minimum lit depth $Z_k$ (the depth of the point $P_k$) and then exhibits a rapid increase until it reaches saturation, while it is smoothly increased with scene depth when ISL is omitted. The saturation indicates that backscatter dependence on the scene point position can be safely omitted after $Z_{sat}$, where the scattered light by the particles becomes negligible (Figure \ref{fBackscatter:Bz:b}): 

\begin{equation}
B(Z)=B(\infty), \forall Z\in [Z_{sat},\infty].
\end{equation}

According to this observation, as soon as the scene is beyond the saturation depth value $Z_{sat}$, we can safely approximate the backscatter component value with the backscatter created when camera looks at infinity $B(Z)\simeq B(\infty)$. Interestingly, we found that even for depths $Z<Z_{sat}$ this approximation is valid as soon as we consider the total measured brightness $E$ on the sensor.

\begin{figure}[htp]%
     \centering
     \subfloat[][Direct \& Backscatter components]{\label{fBackscatter:DB}\includegraphics[height=0.115\textheight]{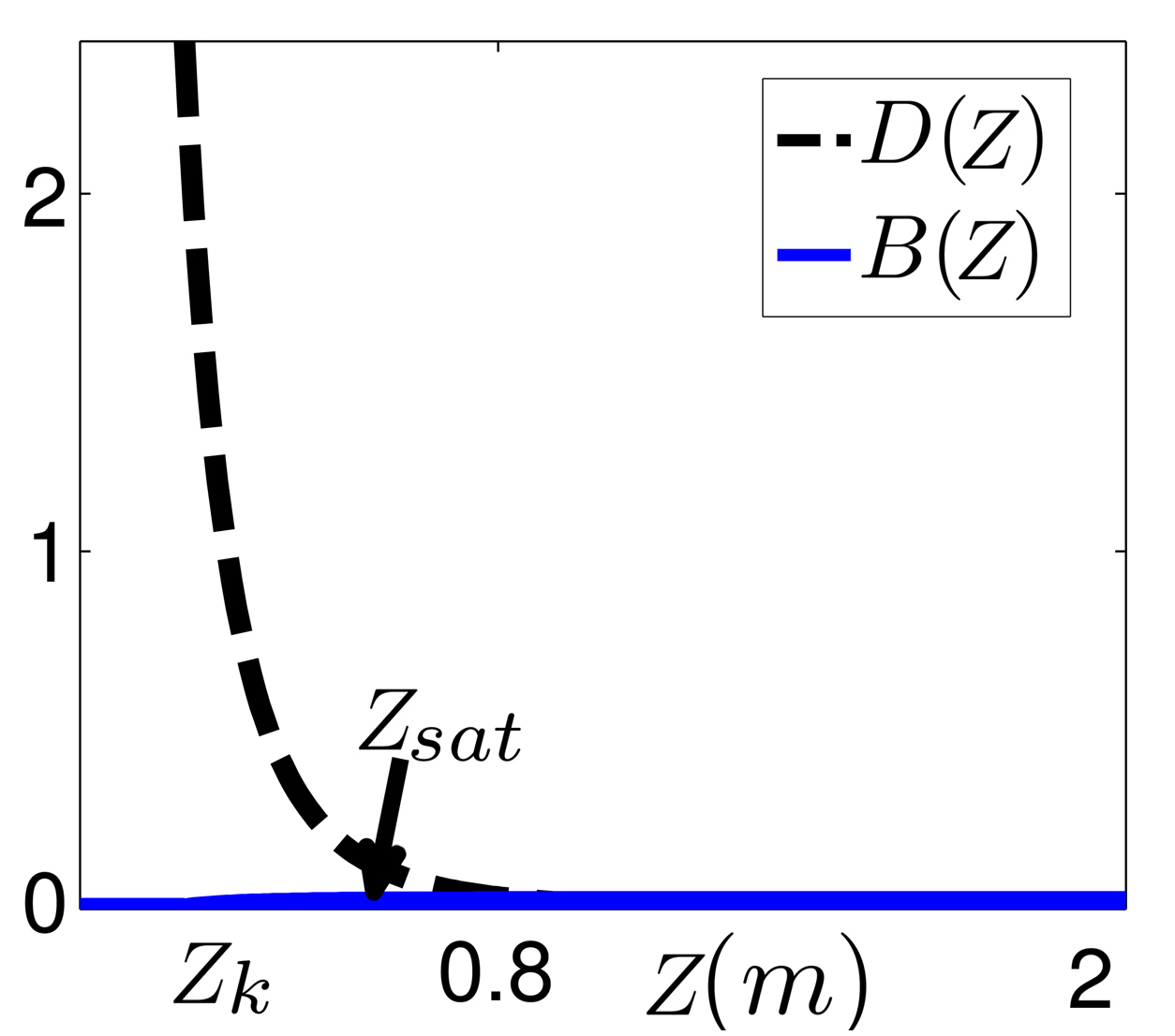}} 
     \hspace{2.5mm}
    \subfloat[][Backscatter \& Total measured brightness ]{\label{fBackscatter:e}\includegraphics[height=0.115\textheight]{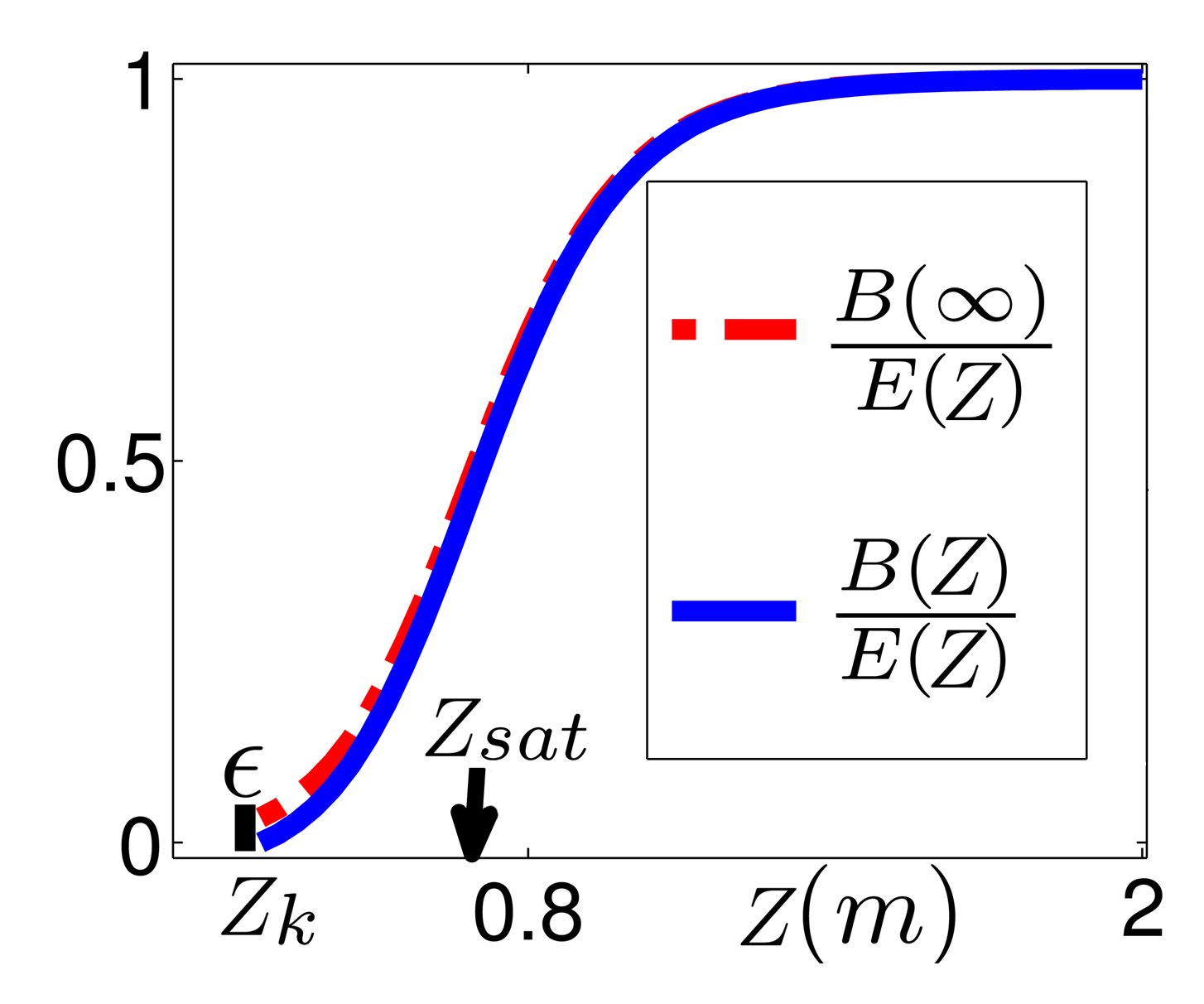}}%
     \vspace{-0mm} 
\caption{In order to define the error of approximating the backscatter at every depth $B(Z)$ with the backscatter from an infinite depth $B(\infty)$, the total measured brightness $E(Z)$ should be considered (Equation \ref{eLight:E}). For scene points after the depth $Z_{sat}$ the error is insignificant due to backscatter saturation. For scene points before the saturation depth the direct component is strong and dominates the dynamic range of the sensor. Thus, the backscatter approximation has again an insignificant impact with respect to the measured brightness. }
    \label{f:Bz}
\end{figure}

If the scene is at small depths below $Z_{sat}$ we expect the incident illumination on the scene and the final direct component $D$ to be high. Figure \ref{fBackscatter:DB} shows the respective direct component over the varying depth $Z$, along with the absolute values of the backscatter component of Figure \ref{fBackscatter:Bz:a}. For small depths where backscatter is un-saturated, the measured brightness seems to be dominated by the direct component intensity. Thus, in order to examine the impact of neglecting the backscatter dependence on depth, we can compare the ratio of the true backscatter with respect to the measured intensity $\frac{B(Z)}{E(Z)}$  with $\frac{B(\infty)}{E(Z)}$ which approximates backscatter with its infinite value regardless of the true scene depth. Figure \ref{fBackscatter:e} indicates that these two differ by a small error value $\epsilon(Z)$ at every depth. This is very small even for depths below $Z_{sat}$ because there the measured intensity is dominated by the direct component and thus the impact of backscatter is insignificant. In this case, backscatter could also be totally neglected with respect to the direct component as it was proposed in \cite{Negahdaripour-PS} (Section \ref{sIm:Phot}). However, as we show in the next sections neglecting backscatter entirely is valid only for a very small depth range compared with our proposed backscatter approximation.

\subsection{Simulations}
\label{sBackscatter:Simulations}

\begin{figure*}[htp]%
     \begin{minipage}[c]{.18\textwidth}
     \centering
     \subfloat[][Pixel position \& Scene depth]{\label{fBackscatter:Var:a}\includegraphics[height=0.22\textheight]{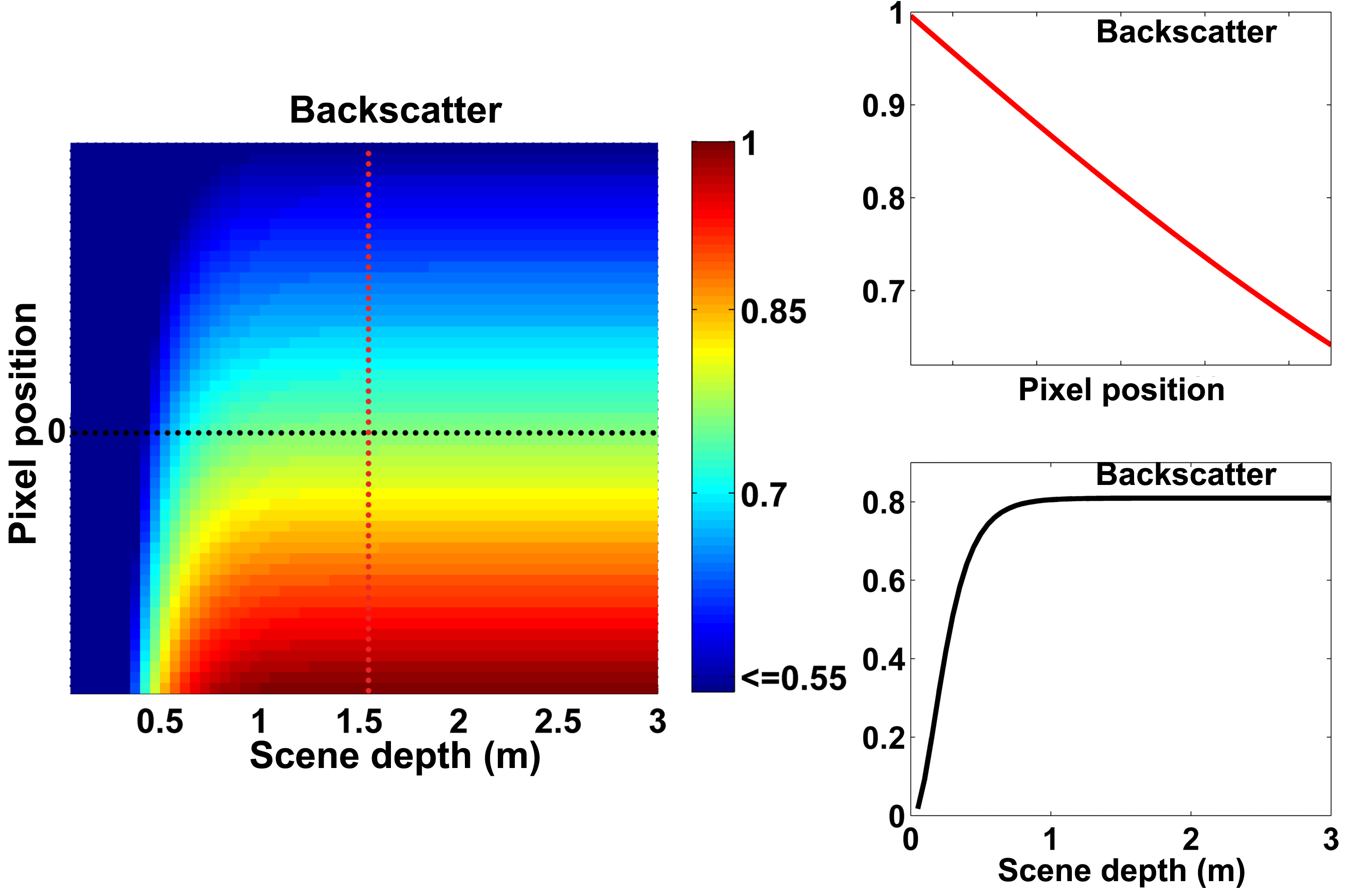}} 
          \end{minipage}
          \hspace{50mm}
               \begin{minipage}[c]{.18\textwidth}
               	     \centering
     \subfloat[][Source position \& Scene depth]{\label{fBackscatter:Var:b}\includegraphics[height=0.16\textheight]{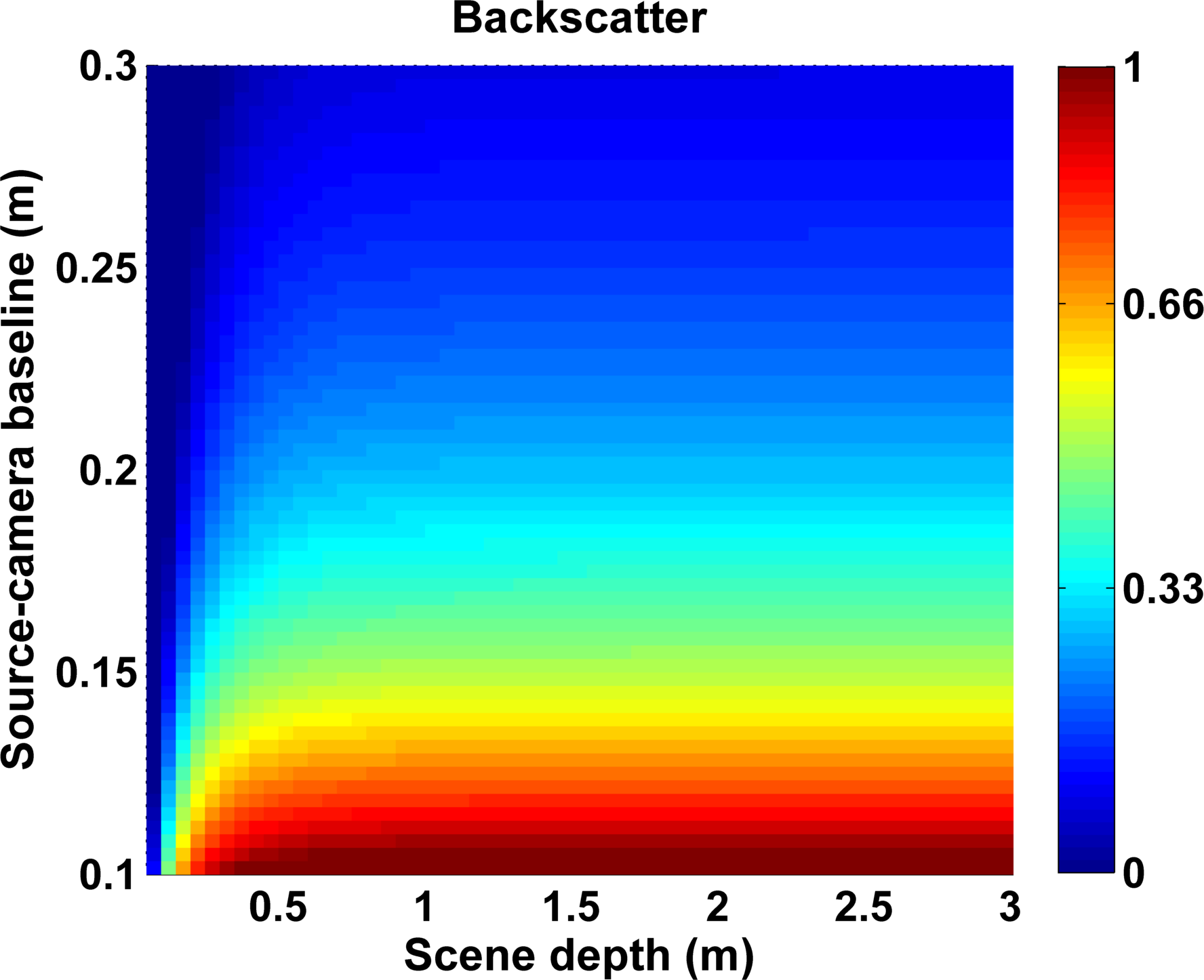}} 
    \end{minipage}
              \hspace{17mm}
     \begin{minipage}[c]{.18\textwidth}
     	     \centering
     \subfloat[][Source beam angle \& Scene depth]{\label{fBackscatter:Var:c}\includegraphics[height=0.16\textheight]{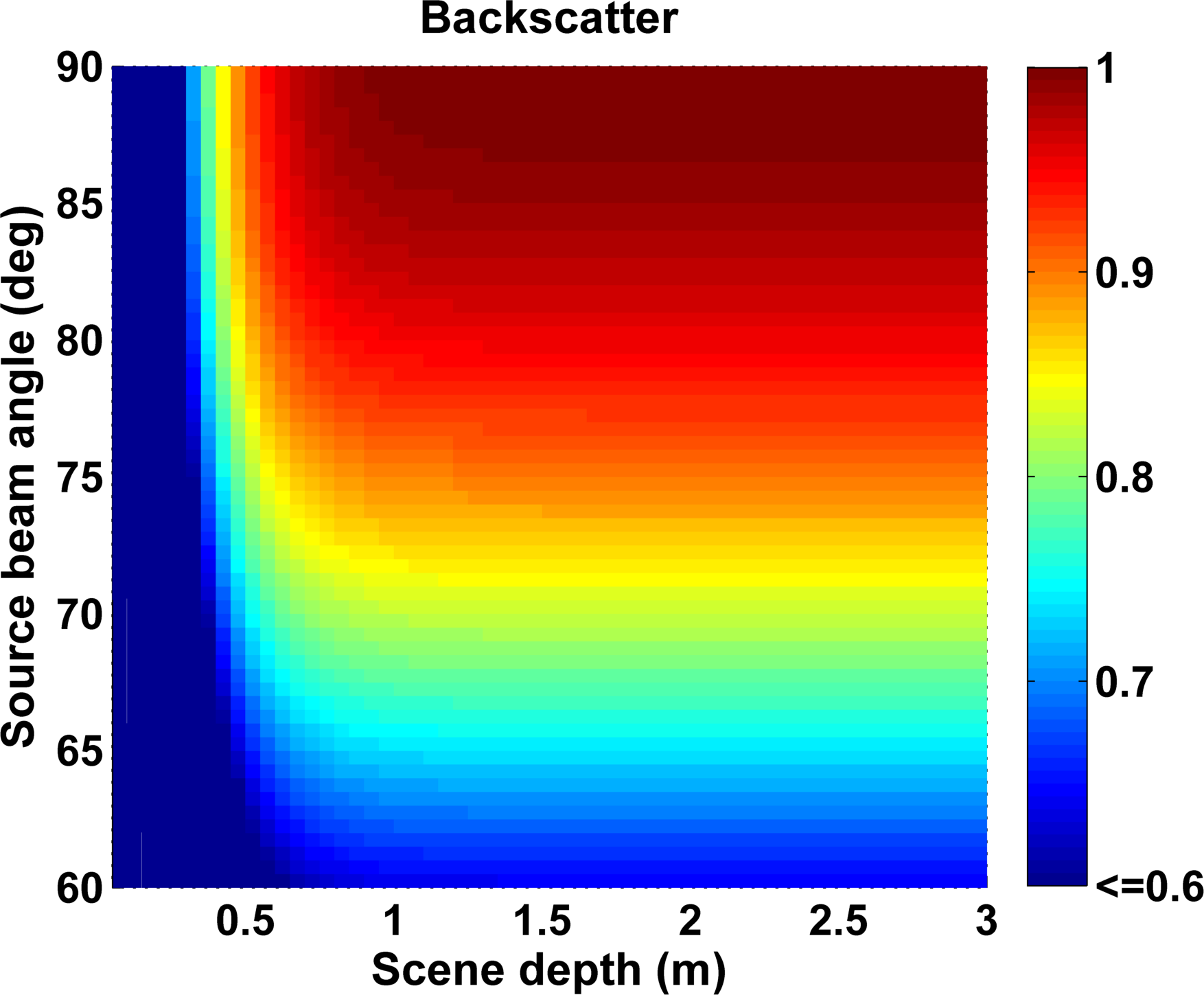}} 
               \end{minipage}
\caption{Backscatter variation with scene depth and (a) pixel position, (b) light source position, and (c) light source beam angle. In all cases it is evident that backscatter for every pixel varies significantly with the pixel position and the light source characteristics but it becomes saturated (varying insignificantly) after a small distance from the camera. }
    \label{fBackscatter:Var}
\end{figure*}

We have run extended numerical simulations in order to evaluate how backscatter varies with the light source characteristics and scene depth. Specifically, we used Equation \ref{eLight:B} and we simulated the backscatter component considering different: (a) scene depths and pixel positions, (b) scene depths and light source positions, and (c) scene depths and light source beam angles. In Figure \ref{fBackscatter:Var} it can be noticed that backscatter varies strongly with changes in pixel position or light source characteristics but it always reaches a saturation value at some depth after which it remains unchanged. Specifically, Figure \ref{fBackscatter:Var:a} shows how backscatter varies for different-adjacent pixel positions on the sensor (for a specific light source position), with respect to variation with scene depth. The graphs in the right correspond to cross sections that indicate this variation more clearly; the red line indicates how backscatter varies along an image column\footnote{We considered a sensor with a small FOV (5 degrees) in order to emphasize the strong backscatter variation on the sensor. For wider FOV this was even stronger.}, assuming that all the pixels image scene points with the same depth. The variation is strong and smooth suggesting that the measured backscatter depends strongly on the pixel position. The black line indicates how backscatter varies for a specific pixel, assuming that this images scene points with varying depth. It can be noticed that backscatter variation is insignificant beyond a certain depth point. 
 
The results are similar considering variations in the other light source characteristics. Figure \ref{fBackscatter:Var:b} indicates how backscatter varies with respect to the light source position (for a specific pixel) and scene depth, and Figure \ref{fBackscatter:Var:c} how it varies with respect to the beam angle of the light source and scene depth. In every case it can be noticed that a small change in light source characteristics has a big impact on the created backscatter, while any change in scene depth is insignificant beyond the small saturation depth. Thus, the backscatter value at every pixel is strongly dependent on the relative position of every pixel with respect to each source and the source's beam angle, but it is independent on the scene depth after a small distance from the camera.

Next, we evaluated the amount of error that is reflected to the Photometric Stereo reconstruction (average error in degrees between the estimated and the ground-truth normal vectors) when our backscatter saturation approximation (approximating the backscatter for every pixel with the backscatter value at infinite depth) is used in different imaging conditions. This was compared with the previous assumptions (Section \ref{sIm:Phot}) that (a) backscatter can be neglected entirely (No Backscatter), (b) backscatter is independent of light source position (Narasimhan et al.), and (c) backscatter is modeled according to Equation \ref{eLight:B} where no approximation is used (original model). Thus the objective of the simulations is to estimate the validity of the different backscatter model approximations compared with the original model which requires full a-priori knowledge of all the unknowns.

We simulated a 16-bit sensor, different levels of additive Gaussian sensor noise, a sphere object, $4$ light sources with different characteristics, and various scattering conditions. We considered realistic scenarios underwater where the reconstruction also suffers from error due to noise. For example, when the distance between the camera and the scene is very large, the direct component is very low and the image is dominated by the backscatter component. In that case, all methods fail (even the original model) since the backscatter compensation leaves only sensor noise and no reconstruction is feasible.

\begin{figure*}[htp]%
 	\centering
 	\includegraphics[width=0.9\textwidth]{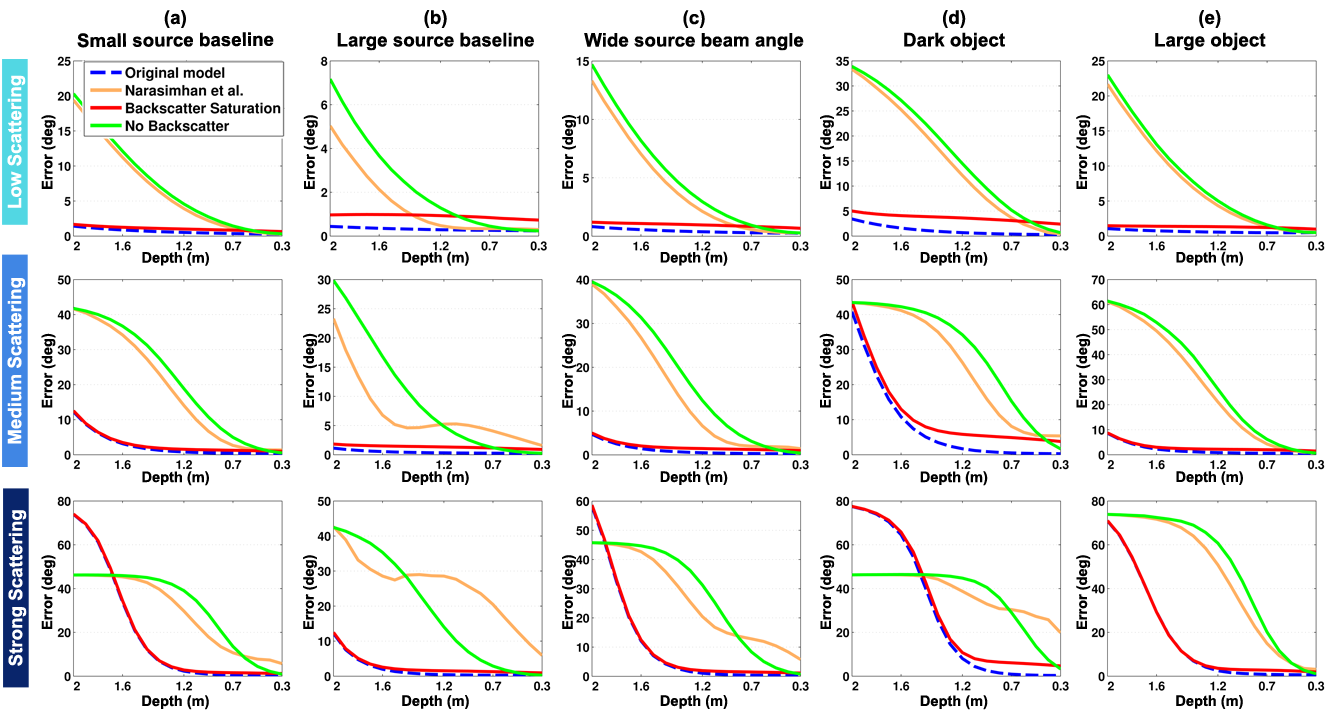}
 	\caption{Simulations for different source, distance, scattering and object characteristics. The reconstruction error between the estimated and the ground-truth normal map of a sphere object is estimated at every depth, considering that: no model approximation is used for the backscatter component (blue dotted line), backscatter can be approximated by its infinite depth value (proposed - red line), backscatter is modeled neglecting its dependence on light source position (orange line), and backscatter is neglected entirely (green line). The x-axis depth values are in decreasing order in order to simulate a robotic platform that gradually approximates the scene starting from far-away (2m) to close-up (0.3m). }
 	\label{fBackscatter:Sim_Each}
\end{figure*}

\begin{figure}[htp]%
	\centering
	\includegraphics[width=0.275\textwidth]{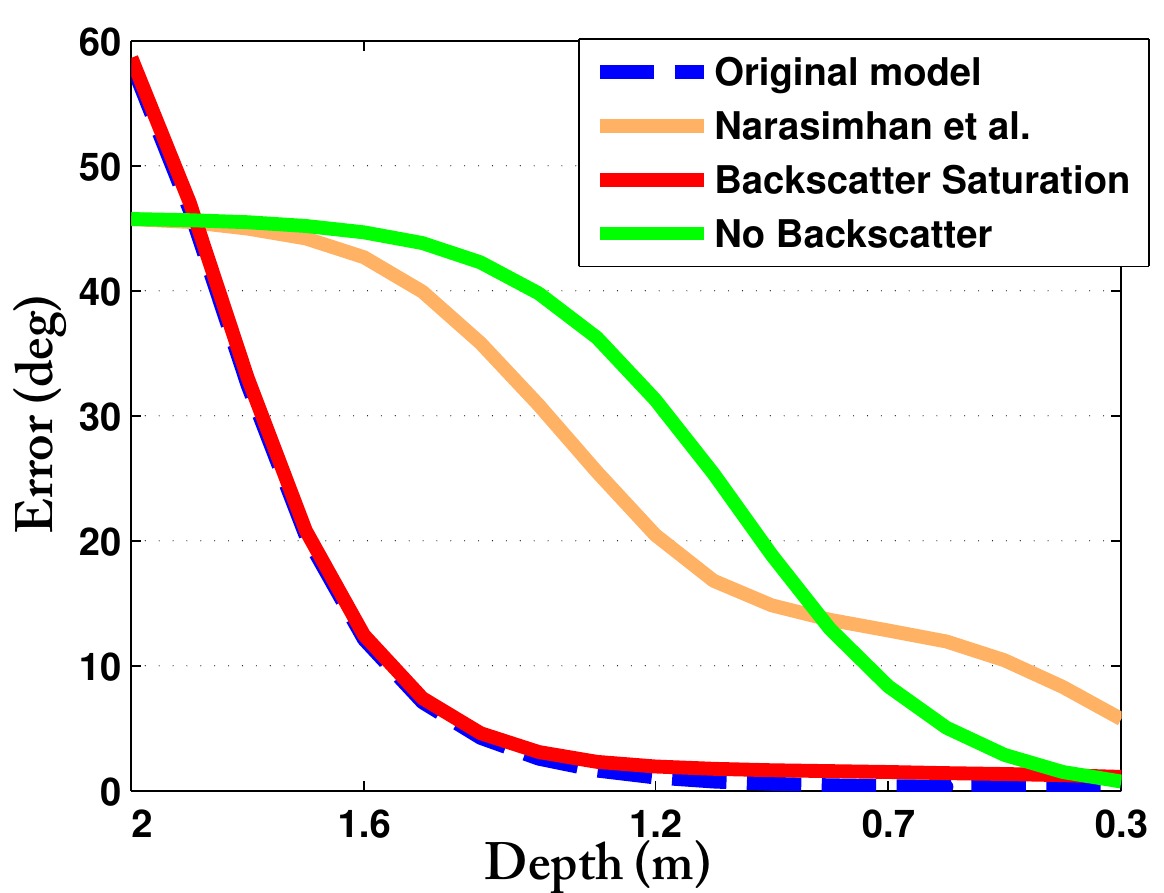}
	\caption{The average error in the estimated normals using Photometric Stereo, when different approximations are used for the backscatter component. Our proposed approximation (red line) assumes that backscatter is depth-independent and causes an insignificant reconstruction error compared to the original model.}
	\label{fBackscatter:Sim_All}
\end{figure}

Figure \ref{fBackscatter:Sim_Each} shows the results for several different cases\footnote{Small and large baselines correspond to camera-source distances of $0.4m$ and $1m$ respectively. Wide source beam angle corresponds to sources with beam angle of $90^o$. Dark object corresponds to albedo equal to $0.2$. Large object corresponds to sphere radius equal to $0.4m$.}. Our proposed backscatter model approximation  that neglects the dependence on scene depth outperforms the previous approximations and yields very small error compared to the original model.  Neglecting backscatter entirely or neglecting inverse-square fall-off reflects significant error to the reconstruction almost in all scattering conditions and scene depths. Neglecting backscatter is more effective than our model only in very small depths. However, even in these cases the error of our model is still small (between $0.5-2$ degrees) and close to the error of the original model.  Figure \ref{fBackscatter:Sim_All} shows the average reconstruction error for all the scenarios\footnote{Source baseline ranging between $0.1-1.5m$ and beam angle between $40-90^o$, object albedo between $0.2-1$ and size between $0.05-1m$, additive gaussian noise with $\sigma$ between $0-0.05$, total attenuation coefficient $c$ between $0-2m^{-1}$, and scene distance between $0.3-2m$.} we considered in our simulations.

\section{Backscatter Estimation}
\label{sBackscatter:Back_estimation}

We showed through simulations that in typical imaging conditions, approximating backscatter for every pixel with its inifinite value comprises an effective approach. In this section we describe that this results in a very simple backscatter estimation method, either via a calibrated or an uncalibrated manner, which is based on the observation that the backscatter component finally varies smoothly for every pixel across the sensor.

\textbf{Previous Work:} The task of estimating the backscatter component when directional sources are employed has drawn limited attention compared with the respective cases of diffuse lighting \cite{He,Sch1}. The work of Mortazavi and Oakley \cite{Mort3,Mort} was the only work we found estimating this directly from the image brightness. The dependence of backscatter on the light characteristics and its saturation with scene depth were omitted in this work and as it was assumed, the measured backscatter was proportional to a low-pass filtered version of the image $B_k\simeq \gamma \bar{E_k}$. $\bar{E_k}$ was used for extracting the illumination variation of the image by suppressing any high-frequency details and for this purpose a recursive Gaussian filter with a large parameter $\sigma$ was used. The parameter $\gamma$ was estimated by minimizing a cost function that was originally described in \cite{OakleyBu}. As we show next, the assumption that backscatter follows a low-pass filtered version of the image is unrealistic in many cases because it is affected by the direct component from the imaged objects. Thus it overestimates the backscatter and introduces high errors to Photometric Stereo.

\textbf{Proposed Calibrated Method:} As it was described in the previous section, for artificial light sources next to the camera backscatter is saturated, and thus the varying integration path that results in an uneven backscatter for every pixel is attributed only to the pixel position with respect to the light source. Thus, the backscatter component can be estimated by capturing images when the camera looks at $\infty$, directly measuring the saturation value $B_k(\infty)$ of every pixel. This calibration step should be done separately for every source $k$, creating a backscatter lookup table for each pixel-source combination. In a finite tank, this can be done using a flat matte black canvas ($\varrho=\|\boldsymbol{n}\|=0)$ to produce $D_k=0$ (Equation \ref{eLight:D}). 

\begin{figure*}[tp]%
	\centering
	\includegraphics[width=0.9\textwidth]{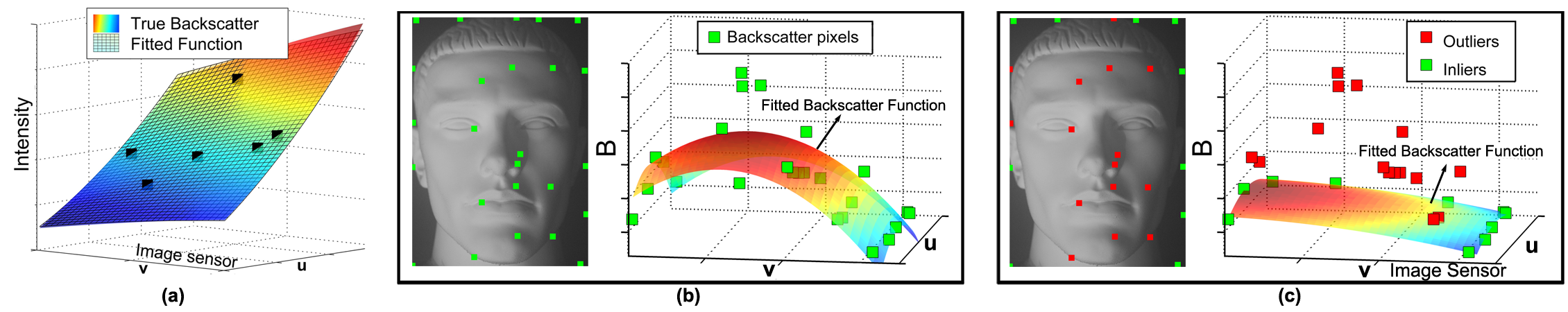}
	\caption{(a) The simulated ground-truth backscatter function is smooth and can be approximated with a quadratic function using only $6$ points. (b) Our proposed backscatter estimation method for a single image. The image is originally divided into blocks and the pixel with the lowest intensity is selected in every block. Naturally, some of the selected pixels will correspond to outliers as their measured brightness consists of an additive non-zero direct component. (c) The outlier pixels are rejected using a modified RANSAC approach and the final backscatter component is estimated by fitting a quadratic function. }
	\label{fBackscatter:BFit}
\end{figure*}

\textbf{Proposed Automatic Method:} As the integration path varies smoothly for every pixel, the respective backscatter function is also smooth. Specifically, $B_k(u,v)$, where $(u,v)$ denotes the pixel position on the sensor, would have its maximum at the pixel position which is closest to the source $k$ and then smoothly decrease for the rest of sensor pixels (Figures \ref{fBackscatter:B_Variation_1} and \ref{fBackscatter:B_Variation_2}). This smoothness gives us insight that knowing the backscatter intensity of only a few pixels, we can approximate the whole smooth backscatter function over the sensor:
\begin{equation}
\label{eBackscatter:f_k}
B_k \simeq f_k(\boldsymbol{\Omega},\boldsymbol{\alpha}),
\end{equation}
where $\boldsymbol{\Omega}$ is the set of all pixel coordinates $(u,v)$, and $\boldsymbol{\alpha}$ are the unknown parameters of the model that approximates $B_k$. Due to the smoothness of the function and its unique maximum on image border, we found that a $2D$ quadratic function $f_k(u,v)=\alpha_0 + \alpha_1 u^2 + \alpha_2 v^2 + \alpha_3 uv + \alpha_4 u + \alpha_5 v$ can estimate accurately the true $B_k$ function of Equation \ref{eLight:B}. A set of at least $6$ points with known backscatter component are needed in order to define the $6$ unknown coefficients of $\boldsymbol{\alpha}$, although more pixels would be necessary for robustness, as we describe in the next section. Figure \ref{fBackscatter:BFit} shows the true backscatter function $B_k$ over the sensor simulated using the original model of Equation \ref{eLight:B}, and the resulting fitted quadratic function $f_k$ using the backscatter values of only $6$ points.

The problem now comes to the selection of at least $6$ backscatter pixels that are the input for our regression. Potential candidates are pixels $(u_B,v_B)$ that correspond either to dark scene points, \emph{i.e.} $\varrho(u_B,v_B)=0$, or to infinite depth points $Z(u_B,v_B)=\infty$ in the object surrounding. From Equation \ref{eLight:D}, the respective direct component for these pixels will be $0$ and hence the measured brightness corresponds to the actual backscatter intensity: $ E_k(u_B,v_B) = \cancelto{0}{D_k(u_B,v_B)} + B_k(u_B,v_B)$. 

In order to select a potential set of backscatter pixels for which $D_k(u_B,v_B)=0$, we divide the image into a number of $N\times N$ blocks  and choose the pixel with the lowest intensity in each block (Figure \ref{fBackscatter:BFit}.b). In reality, not all of the selected points have zero direct component, which introduces a number of outliers. For this purpose, we exploit a RANSAC approach \cite{Forsyth} which iteratively evaluates different $6$-point sets and assigns a score to each set according to how well the quadratic function fits to all the originally selected points after rejecting outliers. We also take advantage of the physical characteristics of our model in order to facilitate the outlier rejection. 


Specifically, given that backscatter for each source has its maximum on a border pixel that is closer to the source, we reject solutions that estimate the maximum of $f_k$ on non-border pixels. Furthermore, in our case the outliers should be always additive to our model $f_k$ since they correspond to a positive direct component: $f_k+D_k,\, D_k>0$. Thus, we penalize solutions that have outliers below the fitted function, by adding the absolute number of these outliers to the RANSAC score count. Figure \ref{fBackscatter:BFit}.c shows the resulting estimated function $f_k$, together with the inliers and outliers of our RANSAC approach. The specific example corresponds to a difficult case since a white object was used and the original image was cropped so that only a small number of backscatter pixels exist in the image. Our proposed method has rejected all the outlier pixels (those that correspond to the object surface) and has estimated the backscatter given only the surrounding pixels. 

This procedure yields an automatic backscatter estimation for each light source, which requires no prior knowledge about the characteristics of the source, the medium or the scene. It is only based on the assumption that at least 6 pixels within the whole image should correspond either to dark points on the object or to infinite depth in the object surrounding.

\section{Real Experiments} 
\label{sBackscatter:Experiments}

We have performed a large number of experiments in order to evaluate the effectiveness of our proposed method. First we show the results from all the experiments in a controlled experimental setup that was manufactured for the purposes of this work, next we demonstrate the results from the Photometric Stereo system we implemented on a real underwater vehicle, and finally we show the potential of our method for single image restoration in murky water.

\subsection{Water Tank Experiments}
\label{fBackscatter:Tank_Experiments}

\begin{figure}[htp]%
       \centering
     \includegraphics[width=0.45\textwidth]{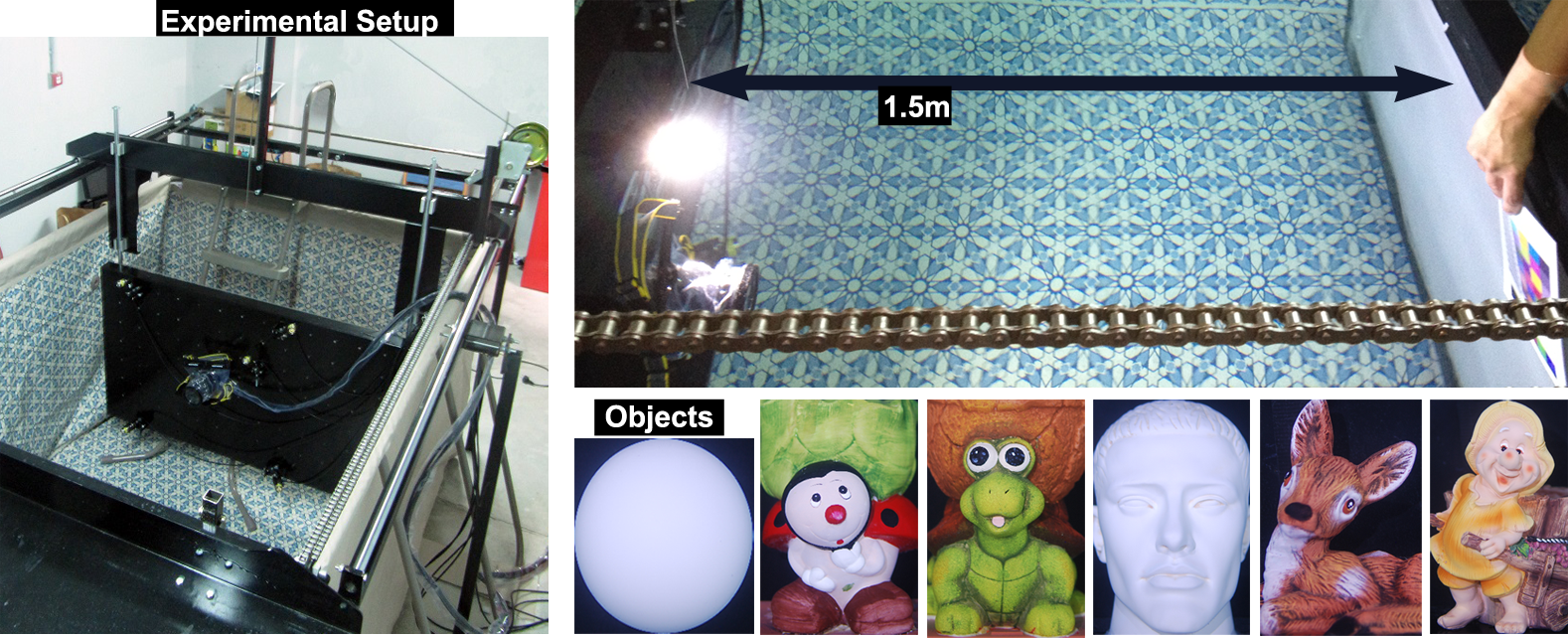} 
        \caption{Our experimental setup consists of a big water tank, a single camera and $4$ light sources that were all immersed into the water. The water murkiness level was varied by diluting increasing amounts of milk into clean water. Several matte objects and a black canvas were imaged for the experiments.}
    \label{fBackscatter:Setup}
\end{figure}

Our experimental setup (Figure \ref{fBackscatter:Setup}) consists of a rectangular-frame pool  with a water volume of roughly $5000L$. Both the underwater lights and the camera were placed in the water, imitating the setup of an underwater robotic platform. Specifically, $4$ lights were on the corners of a square baseline with side length $0.8m$ around the camera. The camera is a Nikon D60 with a AF-S Nikkor $35mm f/1.8G$ lens. The imaged objects are matte, their size (each dimension) is about $10cm$ and they were all captured at $1-1.2m$ depth. To simulate the scattering effect, we made a linear scale of $15$ turbidity steps ranging from totally clean up to heavily murky, by adding milk to the water as in \cite{Narasimhan-Structured,Murez}.

\textbf{Backscatter Saturation \& Estimation:} First, a black canvas was imaged from various camera-scene distances using light sources that were placed at different positions from the camera. In this way, it was evaluated whether the measured/calibrated backscatter component (the measured brightness for every pixel corresponds directly to the backscatter since the canvas was dark) saturated with scene depth. 
 
\begin{figure}[htp]%
      \centering
      \subfloat[][Low scattering (0.15l)]{\label{fBackscatter:B_dis1}\includegraphics[height=0.14\textheight]{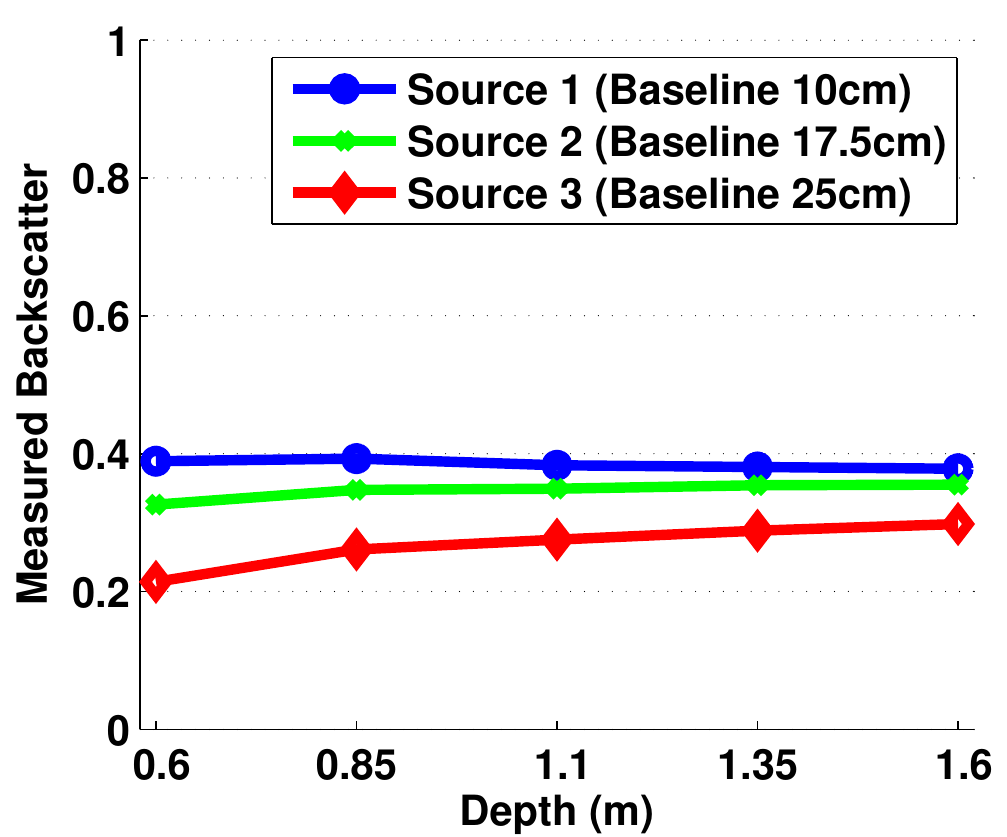}} 
      \hspace{0.05mm}
     \subfloat[][Mid scattering (0.6l)]{\label{fBackscatter:B_dis2}\includegraphics[height=0.14\textheight]{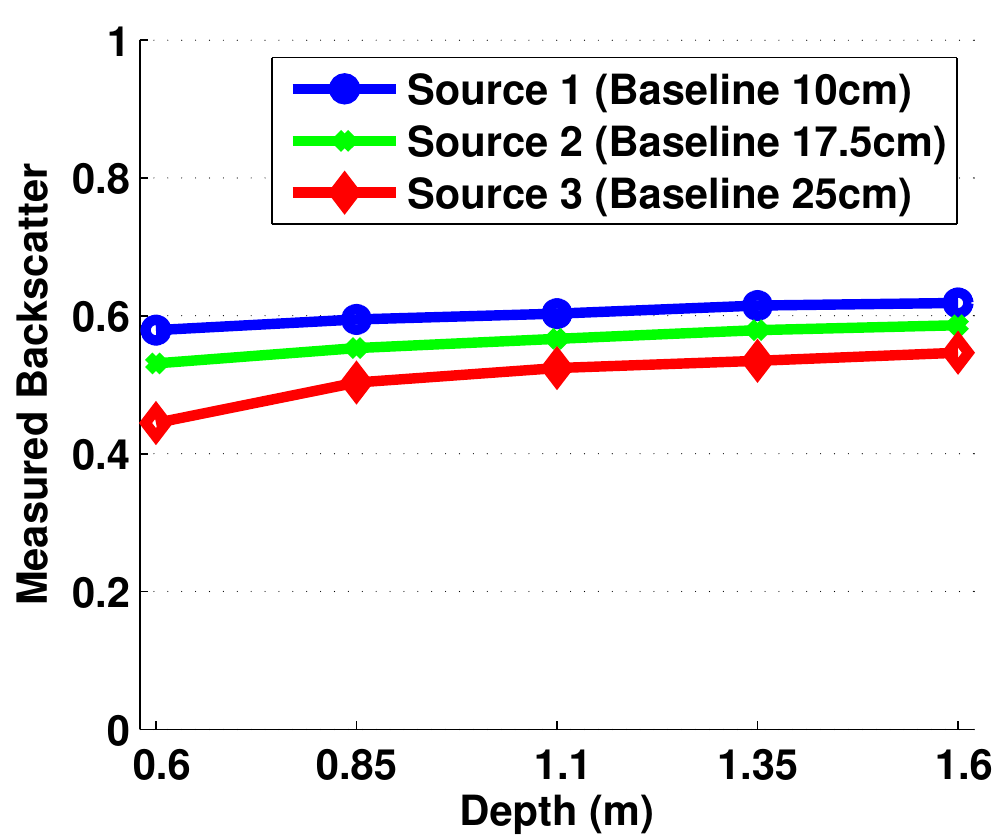}}%
      \vspace{-0mm} 
 \caption{A black canvas was imaged from various distances within our water tank using light sources at different positions from the camera (baselines). In this way, the backcatter component for every pixel was measured directly and its dependence with scene depth was evaluated. In every case, backscatter differed with source position but it was saturated after a small distance, enforcing our analysis for the backscatter component. }
     \label{fBackscatter:B_dis}
 \end{figure}

Figure \ref{fBackscatter:B_dis} shows the measured backscatter component for different distances, light source positions and scattering conditions. Backscatter changed evidently with a small displacement in the source position but in every case it changed insignificantly with scene depth after a small distance from the camera. Thus our experimental results coincide with the simulations of Section \ref{sBackscatter:BVariation}.

\begin{figure}[tp]
  \centering
 \subfloat[][]{\label{fBackscatter:Back_RMSE:a}\includegraphics[height=0.15\textheight]{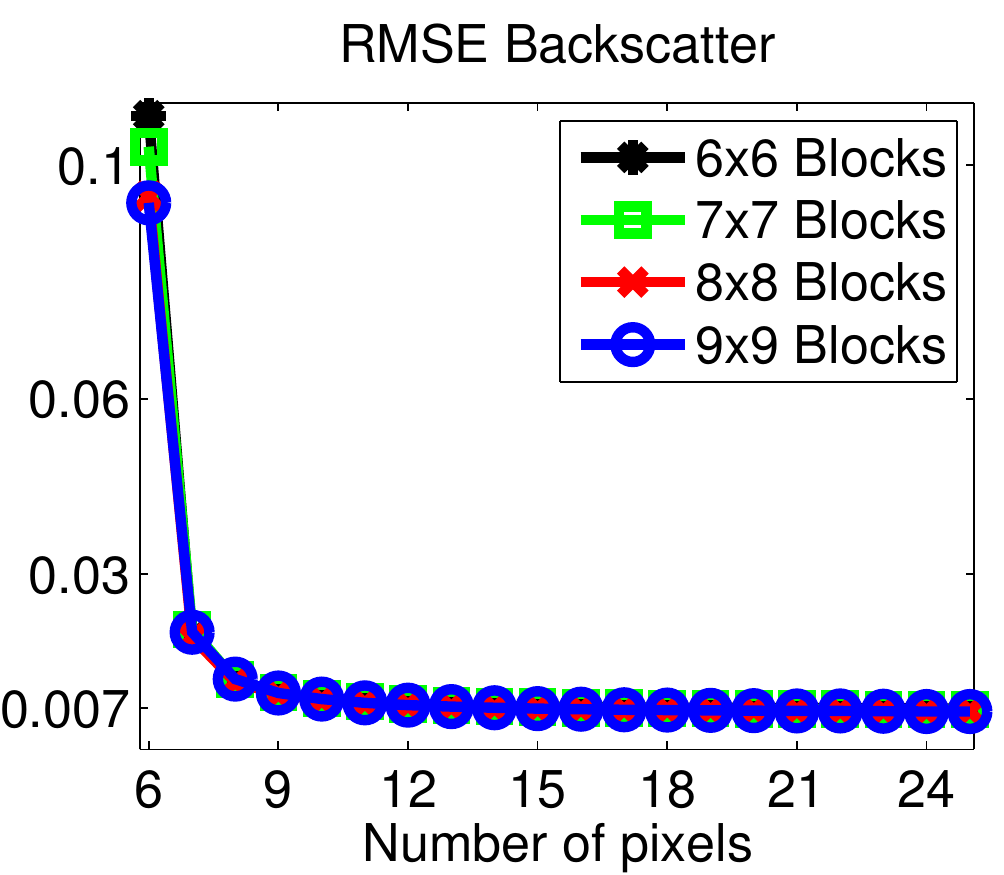}}%
 \hspace{0.7mm} 
 \subfloat[][]{\label{fBackscatter:Back_RMSE:b}\includegraphics[height=0.15\textheight]{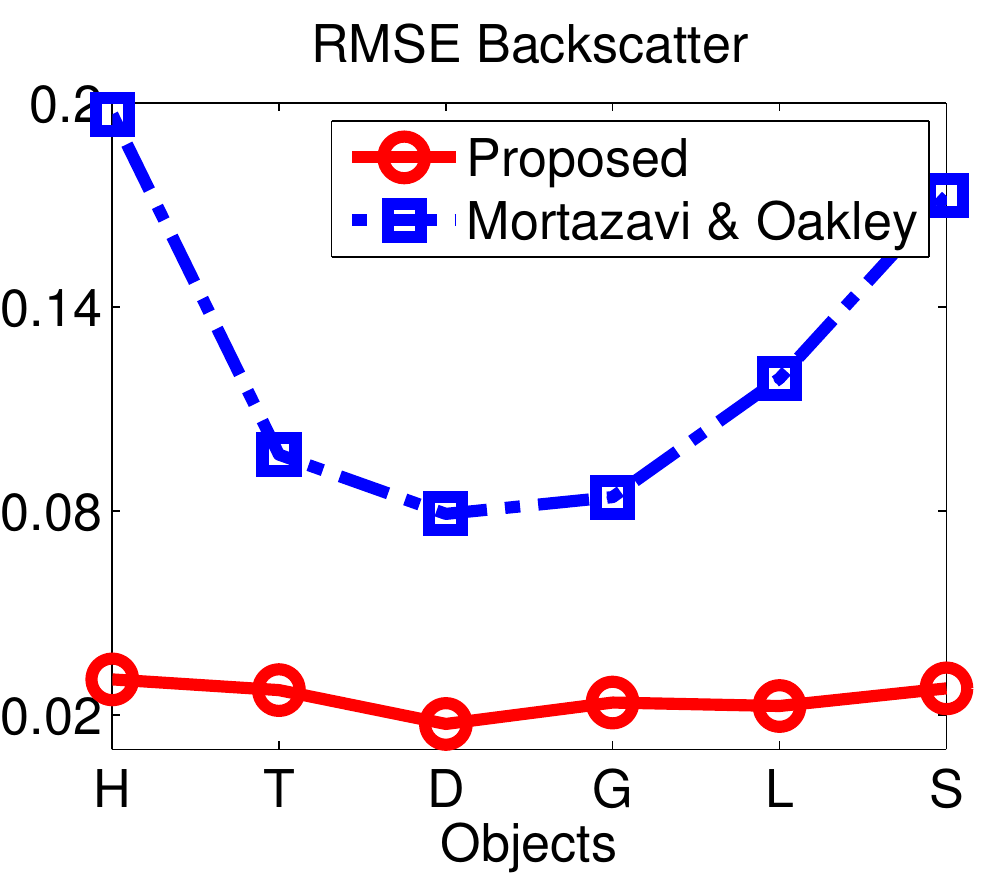}} 
   \caption{ \label{fBackscatter:Back_RMSE} (a) Backscatter estimation error according to the number of backscatter pixels (x-axis). Regardless of the number of image blocks, the backscatter can be approximated effectively by a quadratic function using at least $8$ pixels. (b) Backscatter estimation error for the Head, Turtle, Deer, Gnome, Ladybug and Sphere objects (x-axis). Our method outperforms the method of Mortazavi and Oakley \cite{Mort,Mort3}.}.
 \end{figure}

Next, in order to examine how well the quadratic function can fit to the ground-truth backscatter given only a small number of its points, we estimated the $RMSE$ between the real and the estimated backscatter (using Equation \ref{eBackscatter:f_k})  after selecting a different random combination of points. As Figure \ref{fBackscatter:Back_RMSE:a} indicates, regardless of the number of the image blocks, the error was as low as noise variation when at least $8$ pixels were used, supporting the validity of the quadratic function selection.

 Then the performance of our automatic backscatter estimation method was compared with the method of Mortazavi and Oakley \cite{Mort3} (Section \ref{sBackscatter:Back_estimation}). Specifically, the estimated backscatter using each method for different objects was compared with the estimated backscatter using the calibrated method that measures the backscatter using a black canvas. Figure \ref{fBackscatter:Back_RMSE:b} shows the estimated RMSE.  Our method outperformed \cite{Mort3} for all the imaged objects. Regarding the Sphere and Head objects, whose images include a significantly large number of white pixels, backscatter was still estimated effectively, while the error for \cite{Mort3} increased significantly, overestimating the backscatter due to the unrealistic assumption that this is proportional to a low-pass filtered version of the image. For these objects our method rejected all the outliers on the white object, approximating the backscatter from a small number of scene points on the background. An advantage of our method is that due to backscatter saturation with scene depth, the selected backscatter pixels do not have to be dark patches on the object; they can also be dark or infinity points on the object surrounding. Thus, our method does not rely on strong assumptions about the scene, such as the diffuse lighting method of \cite{He} which assumed that a dark point exists in a small neighbourhood around every scene patch. In the case where white objects covered the whole image would lead to erroneous backscatter estimation. However, such a case would be rare in deep-sea scenarios where infinite depth usually surrounds the imaged objects, and even then additional frames could be employed by moving the camera to target surrounding dark or infinity pixels. 

 \textbf{Shape Recovery:} Our Photometric Stereo approach was the following: given the original murky-water images, we estimated the uneven backscatter in every image using either our calibrated or uncalibrated backscatter estimation method. Then, the backscatter component was subtracted from the captured images and the remaining direct component was used to estimate the normal vector and albedo for every pixel assuming distant-lighting on the object (Section \ref{sIm:Phot}). The processing time was approximately $2$ seconds for $800\times600$ pixels images, using an Intel Core i5-2410M CPU @ 2.3GHz and a MATLAB implementation.

\begin{figure}[tp]%
       \centering
      \subfloat[][Sphere Object]{\label{fBackscatter:Sphere:a}\includegraphics[width=0.24\textwidth]{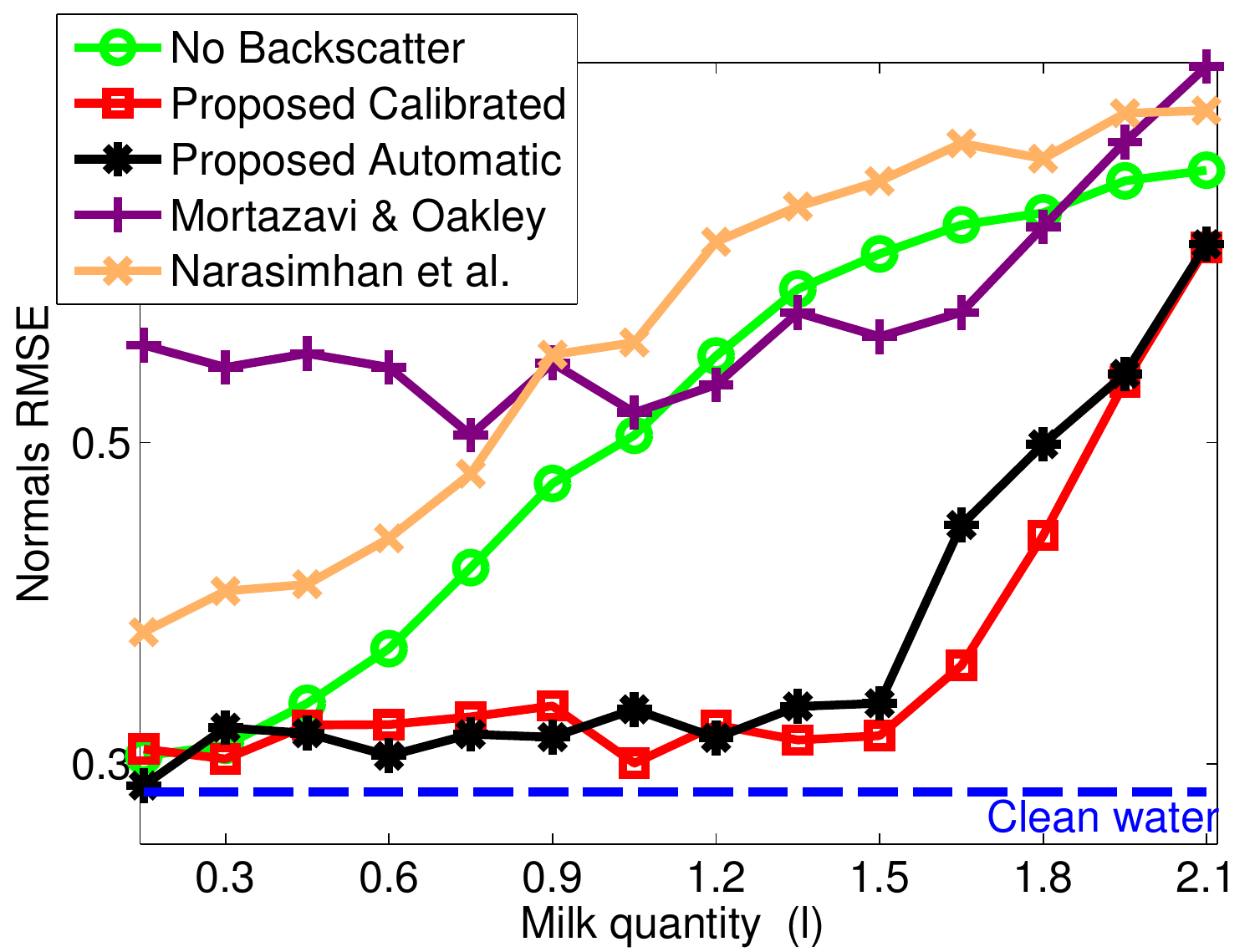}}%
      \subfloat[][Man-made objects]{\label{fBackscatter:Sphere:b}\includegraphics[width=0.24\textwidth]{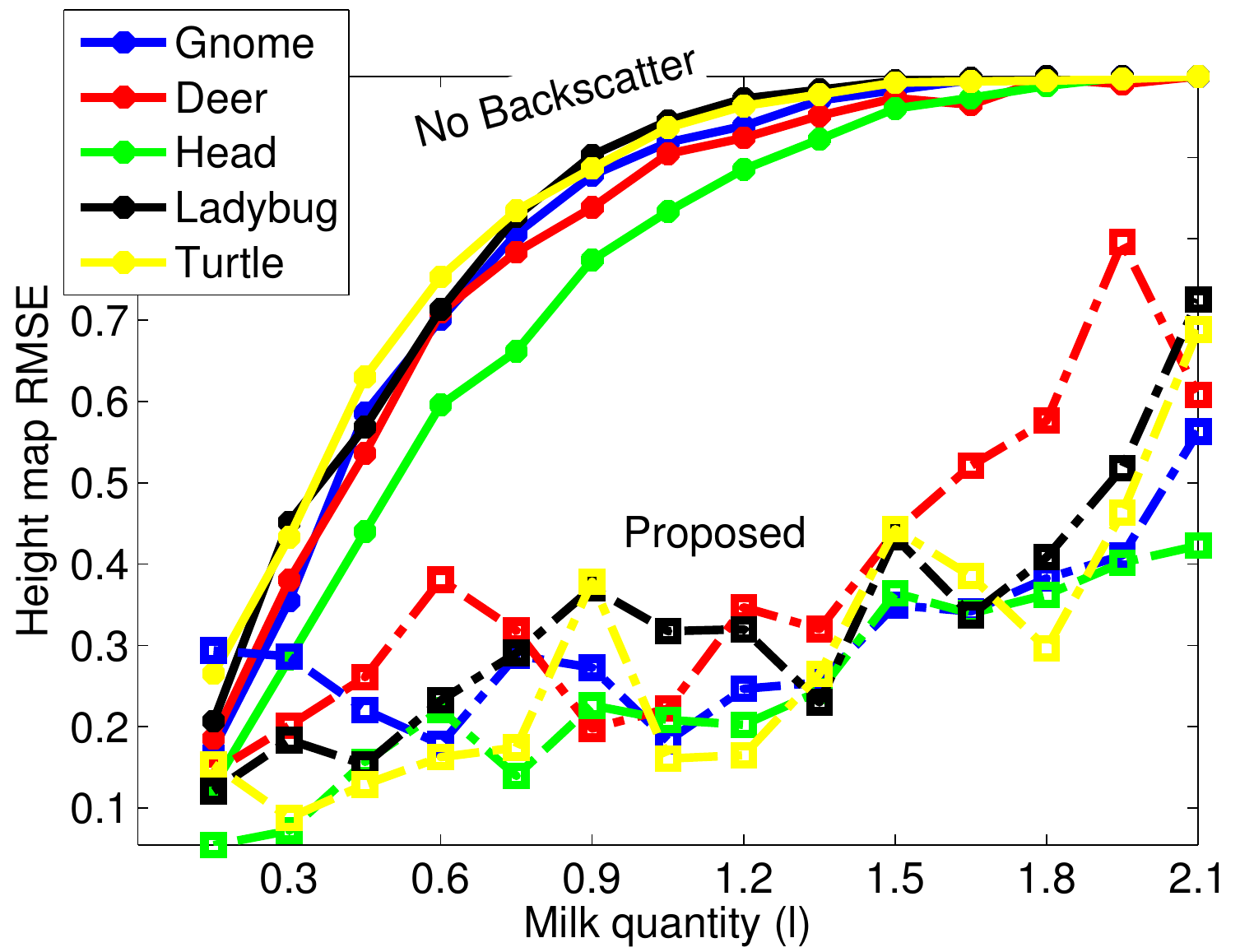}} 
        \caption{Shape reconstruction error using different models for the backscatter component. In (a) the RMSE was calculated between the known ground-truth normals of the sphere and the estimated normals using each method. In (b) the RMSE was calculated between the estimated shape of the objects in clean water and the estimated shape in every scattering level. Our method outperformed all compared approaches.}%
    \label{fBackscatter:Sphere}
\end{figure}

 
The recovered normal vectors at the various water murkiness levels were quantitatively assessed using the Sphere object whose normals are a priori known. Figure \ref{fBackscatter:Sphere:a} shows the $RMSE$ value between the estimated and the ground-truth maps for each scattering level using: our proposed backscatter approximation method (both calibrated and uncalibrated), the method of \cite{Negahdaripour-PS} where backscatter is neglected, the method of Narasimhan et al. \cite{Narasimhan-Structured} where pairs of images are subtracted to eliminate backscatter assuming that this is independent of light source position (Section \ref{sIm:Phot}), and applying Photometric Stereo after estimating and subtracting backscatter using the method of Mortazavi and Oakley \cite{Mort3,Mort} (Section \ref{sBackscatter:Back_estimation}). Our approach yields effective normals estimation, similar to the result in totally clean water for a very wide range of scattering levels. The performance is decreased beyond $1.5 l$ of milk, in a similar manner that other methods are decreased within the lowest murkiness levels. This effect is reasonable, since beyond that scattering level the degrading effects are so severe that the backscatter takes up almost all of the dynamic range of the sensor.

\begin{figure*}[htp]%
	\centering
	\includegraphics[width=0.9\textwidth]{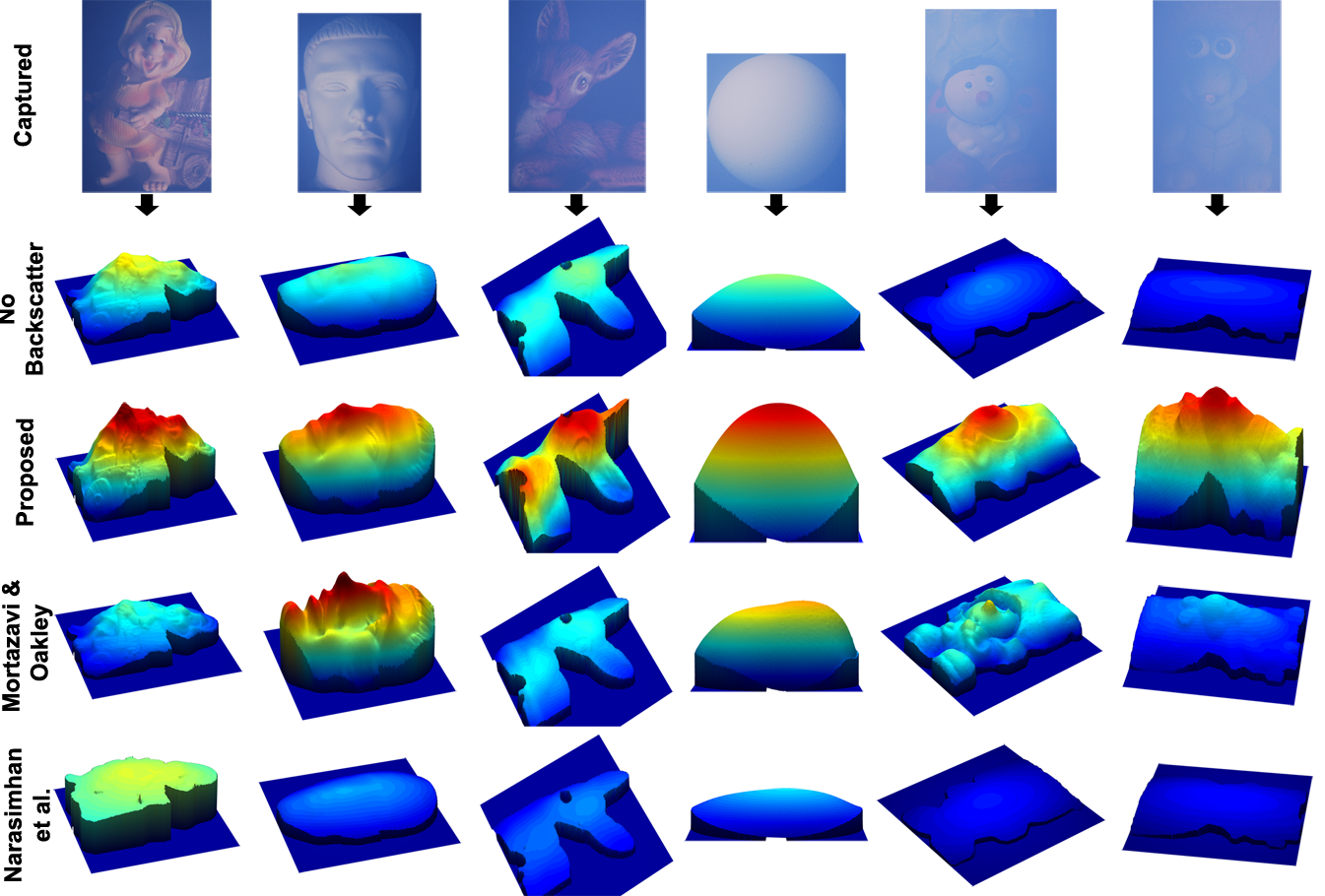}
	\caption{Comparison of the recovered shape using different approaches. The first row shows one of the captured images for every object at a different scattering level. It can be noticed that our proposed approach outperforms all the others for all levels of water murkiness. }
	\label{fBackscatter:Res_Shape}
\end{figure*}

The shape reconstruction results using the man-made objects of Figure \ref{fBackscatter:Setup} were evaluated next. The outputs of the Photometric Stereo method are the normals and the albedo of each pixel. In order to reconstruct the height map from the respective normals we employ the integration method of \cite{Frankot}. Figure \ref{fBackscatter:Sphere:b} shows the $RMSE$  between the reconstructed height of each object in clean water and that estimated at each scattering level, using our proposed method and that of neglecting backscatter which had the best performance amongst the other methods, while Figure \ref{fBackscatter:Res_Shape} compares optically the recovered shape of various objects using all methods. As can be observed, our method successfully preserves the reconstructed shape, while the rest of the methods tend to smoothly flatten the result over increased turbidity levels.

\begin{figure*}[htp]%
	\centering
	\includegraphics[width=0.8\textwidth]{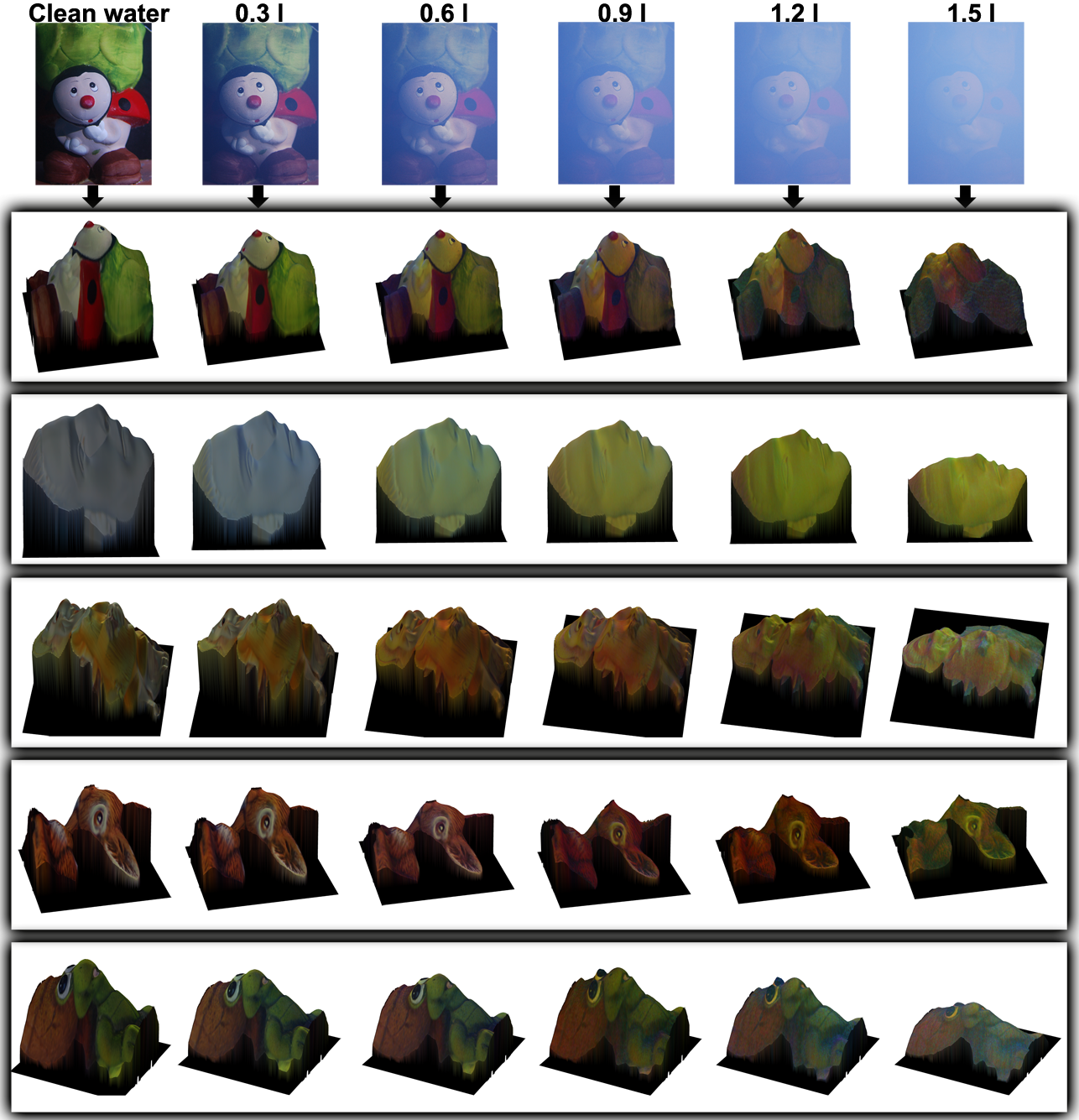}
	\caption{Results of our Photometric Stereo approach in a water tank with increased murkiness condition (indicated by the amount of milk in $l$). The first row indicates the gradual loss of visibility. Shape and albedo are recovered in a dense and detailed manner, even for very high levels of water murkiness.}
	\label{fBackscatter:Res_Final}
\end{figure*}

Figure \ref{fBackscatter:Res_Final} demonstrates our results for various objects and murkiness levels. It can be noticed that the  albedo of the object is restored and the shape reconstruction is dense and detailed even for very high levels of water turbidity. Small error in the low-frequency shape of the objects is present in some cases. This is a well-known drawback of Photometric Stereo that is attributed to the distant-lighting assumption that neglects the low-frequency variation of the scene depth. As we describe in Sections \ref{sIm:Phot} and \ref{sBackscatter:Conclusions}, further refinement can be applied to correct this effect which is beyond the scope of this work. A video with a lot of results and optical comparisons between our proposed method and previous approaches can be found in \url{https://www.youtube.com/watch?v=yUtM9sZBPwk}.


\subsection{Port Water Experiments}
\label{sBackscatter:Port_Experiments}

\begin{figure*}[htp]%
	\centering
	\includegraphics[width=0.92\textwidth]{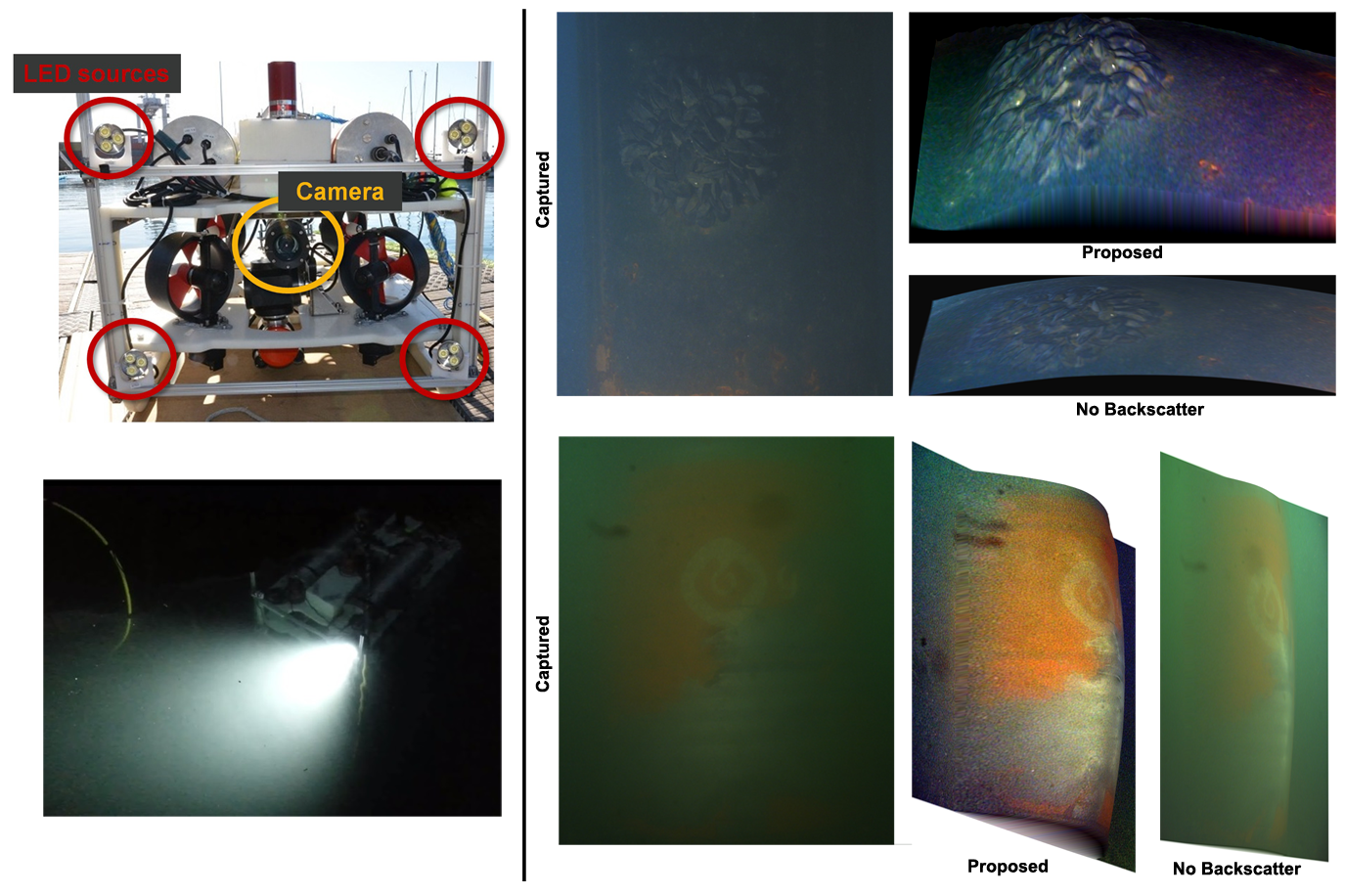}
	\caption{The results of our Photometric Stereo approach in real port water, in Porto, Portugal. Despite the strong scattering condition, our proposed method can yield detailed reconstruction of the imaged targets. Neglecting backscatter yields flat results. Notice the dense high-frequency detail of the shells on the pile object.}
	\label{fBackscatter:Res_Port}
\end{figure*}

A Photometric Stereo system was also implemented on a remotely operated vehicle and was tested in real murky water in the port water of Leix\~{o}es in Porto, Portugal. Figure \ref{fBackscatter:Res_Port} shows the imaging setup. A Lumenera Le-165 camera with a Tamron 219-HB lens were mounted in the middle of the vehicle and four LED sources were mounted in the corners. The camera and the lights were synchronized with each other so that each frame is captured with only one of the light sources on. In order to compensate for any additive light components from the environment, a fifth frame was also captured having all light sources off and it was subtracted from the rest of the frames. 


Figure \ref{fBackscatter:Res_Port} shows the reconstruction results for two different targets that were captured during two separate imaging sessions. One of the original murky water images, the result using our proposed backscatter approximation method, and the result neglecting backscatter can be seen in every case. Our framework recovers a detailed representation of the object surface and albedo, while neglecting backscatter leads to flat estimates. The dense high-frequency detail of the shells on the pile surface (first object) indicates the great potential of Photometric Stereo reconstruction in murky water.

\subsection{Image Restoration}
\label{fBackscatter:Image_Restoration}

\begin{figure*}[htp]%
       \centering
\includegraphics[width=1\textwidth]{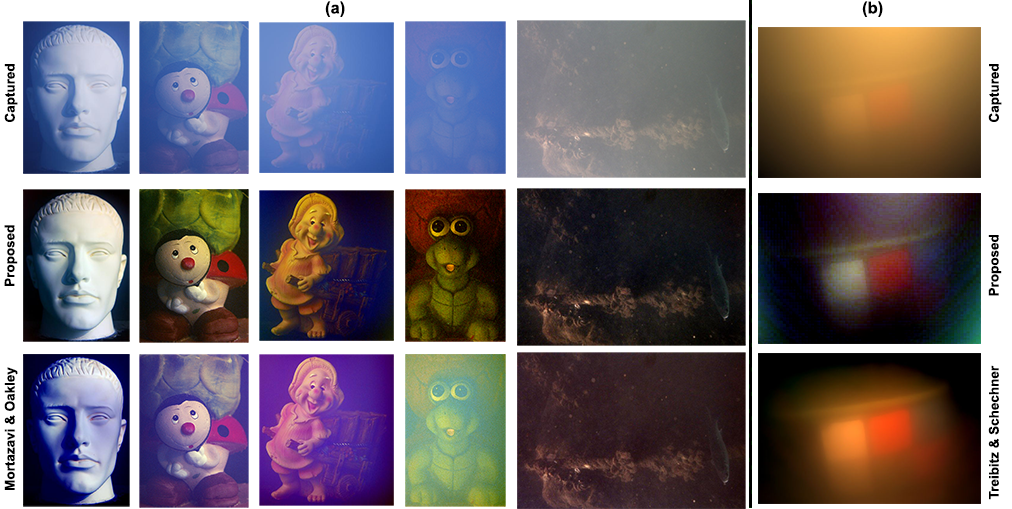}%
      \caption{(a) The results of our backscatter estimation method compared with the method of Mortazavi \& Oakley \cite{Mort}. The first four images correspond to water tank experiments at different scattering conditions and the final one to an experiment in real port water. (b) The restoration result of our method compared with the method of Treibitz and Schechner \cite{Treibitz-Active} where polarizers are used on the light source and camera. The murky water image was adapted from http://webee.technion.ac.il/Sites/People/yoav/research/active-descattering.html. }%
    \label{fBackscatter:Restoration}
\end{figure*}

Finally, we used our proposed backscatter compensation method on single images to demonstrate the potential of our approach for image restoration. In this case, after backscatter was subtracted from the original images the remaining attenuated direct signal was rescaled to recover full contrast. Figure \ref{fBackscatter:Restoration} shows the recovered visibility using our proposed method and the method of Mortazavi \& Oakley \cite{Mort,Mort3}, as well as the method of \cite{Treibitz-Active} where polarizing filters are used. Our approach restores the image contrast and the object albedo even for levels of high water murkiness. 

%
%
\section{Discussion and Conclusions}
\label{sBackscatter:Conclusions}

In this work we showed that the backscatter component can be effectively approximated in murky water relaxing the previous limiting assumptions about the medium (no backscatter) or the imaging setup (light sources and camera outside water). Specifically, we described that the backscatter becomes saturated with scene depth, but it varies significantly across the sensor according to the position of each pixel with respect to the light source.  Since this variation is smooth, the backscatter component can be approximated by a smooth function such as a quadratic function, or be measured directly by capturing images of camera looking at infinity. 

Estimating the backscatter component for every pixel and source directly from the measured images offers significant potentials. The under-determined system of equations for Photometric Stereo can be optimized for orientation and albedo, as the complex backscatter term is removed. As we showed, this yields detailed reconstructions even for high levels of water murkiness. At the same time, single image restoration can be achieved without knowledge about the medium or imaging characteristics. 

Further effects can degrade the performance of Photometric Stereo in murky water. Specifically, errors in the low-frequency representation of the estimated shape can appear when the direct component lighting is assumed to be distant and the object is large compared to the camera-scene distance. In this case, the iterative near-lighting algorithm of \cite{Kolagani, Negahdaripour-PS} can be used instead. Forward-scattering \cite{Murez} can also be modeled to account for resolution loss. Our work focused on the effect and the compensation for the backscatter component which is the strongest degrading factor underwater \cite{Jagger,Mort,Sch1,Schechner-clear,Negahdaripour-Shading} and is complementary with further modeling regarding the direct component.

\ifCLASSOPTIONcompsoc
  \section*{Acknowledgments}
  This work was supported by the
  contract 270180 of the European Communities FP7
  (NOPTILUS). We are very grateful to Underwater Systems and Technologies Laboratory (LSTS) and OceanScan in Porto for the access to the ROV and the help with the port-water experiments.
\else
  \section*{Acknowledgment}
\fi





%

\bibliographystyle{IEEEtran}
\bibliography{egbib}

\begin{thebibliography}{10}
\providecommand{\url}[1]{#1}
\csname url@samestyle\endcsname
\providecommand{\newblock}{\relax}
\providecommand{\bibinfo}[2]{#2}
\providecommand{\BIBentrySTDinterwordspacing}{\spaceskip=0pt\relax}
\providecommand{\BIBentryALTinterwordstretchfactor}{4}
\providecommand{\BIBentryALTinterwordspacing}{\spaceskip=\fontdimen2\font plus
\BIBentryALTinterwordstretchfactor\fontdimen3\font minus
  \fontdimen4\font\relax}
\providecommand{\BIBforeignlanguage}[2]{{%
\expandafter\ifx\csname l@#1\endcsname\relax
\typeout{** WARNING: IEEEtran.bst: No hyphenation pattern has been}%
\typeout{** loaded for the language `#1'. Using the pattern for}%
\typeout{** the default language instead.}%
\else
\language=\csname l@#1\endcsname
\fi
#2}}
\providecommand{\BIBdecl}{\relax}
\BIBdecl

\bibitem{Ortiz-Particle}
A.~Ortiz, J.~Antich, and G.~Oliver, ``A particle filter-based approach for
  tracking undersea narrow telecommunication cables,'' \emph{Machine Vision and
  Applications}, vol.~22, no.~2, pp. 283--302, 2011.

\bibitem{Eustice-Titanic}
R.~M. Eustice, H.~Singh, J.~J. Leonard, and M.~R. Walter, ``Visually mapping
  the rms titanic: Conservative covariance estimates for slam information
  filters,'' \emph{The international journal of robotics research}, vol.~25,
  no.~12, pp. 1223--1242, 2006.

\bibitem{Treibitz-Corals}
T.~Treibitz, B.~P. Neal, D.~I. Kline, O.~Beijbom, P.~L. Roberts, B.~G.
  Mitchell, and D.~Kriegman, ``Wide field-of-view daytime fluorescence imaging
  of coral reefs,'' in \emph{IEEE Oceans}, 2013.

\bibitem{Sarafraz2}
A.~Sarafraz, S.~Negahdaripour, and Y.~Schechner, ``Performance assessment in
  solving the correspondence problem in underwater stereo imagery,'' in
  \emph{OCEANS 2010}, sept. 2010, pp. 1 --7.

\bibitem{Negahdaripour-PS}
S.~Negahdaripour, H.~Zhang, and X.~Han, ``Investigation of photometric stereo
  method for 3-d shape recovery from underwater imagery,'' in \emph{OCEANS'02
  MTS/IEEE}, vol.~2, 2002, pp. 1010--1017.

\bibitem{Negahdaripour-Shading}
S.~Zhang and S.~Negahdaripour, ``3-d shape recovery of planar and curved
  surfaces from shading cues in underwater images,'' \emph{Oceanic Engineering,
  IEEE Journal of}, vol.~27, no.~1, pp. 100--116, 2002.

\bibitem{Narasimhan-Structured}
S.~Narasimhan, S.~Nayar, B.~Sun, and S.~Koppal, ``Structured light in
  scattering media,'' in \emph{Computer Vision, 2005. 2005. Tenth IEEE
  International Conference on}, vol.~1, 2005, pp. 420--427 Vol. 1.

\bibitem{He}
K.~He, J.~Sun, and X.~Tang, ``Single image haze removal using dark channel
  prior.'' \emph{Pattern Analysis and Machine Intelligence, IEEE Transactions
  on}, vol.~99, no.~1, pp. 1956--1963, 2010.

\bibitem{Nay2}
S.~Narasimhan and S.~Nayar, ``Contrast restoration of weather degraded
  images,'' \emph{Pattern Analysis and Machine Intelligence, IEEE Transactions
  on}, vol.~25, no.~6, pp. 713--724, 2003.

\bibitem{Tarel}
J.~Tarel and N.~Hautiere, ``Fast visibility restoration from a single color or
  gray level image,'' in \emph{Computer Vision, 2009 IEEE 12th International
  Conference on}, 29 2009-Oct. 2, pp. 2201--2208.

\bibitem{Chiang}
J.~Chiang and Y.-C. Chen, ``Underwater image enhancement by wavelength
  compensation and dehazing,'' \emph{Image Processing, IEEE Transactions on},
  vol.~21, no.~4, pp. 1756--1769, April 2012.

\bibitem{Schechner-clear}
Y.~Y. Schechner and N.~Karpel, ``Clear underwater vision,'' in \emph{Computer
  Vision and Pattern Recognition, 2004. CVPR 2004. Proceedings of the 2004 IEEE
  Computer Society Conference on}, vol.~1.\hskip 1em plus 0.5em minus
  0.4em\relax IEEE, 2004, pp. I--536.

\bibitem{Jaffe}
J.~S. Jaffe, ``Computer modeling and the design of optimal underwater imaging
  systems,'' \emph{Oceanic Engineering, IEEE Journal of}, vol.~15, no.~2, pp.
  101--111, 1990.

\bibitem{Treibitz-Fusion}
T.~Treibitz and Y.~Schechner, ``Turbid scene enhancement using
  multi-directional illumination fusion,'' \emph{Image Processing, IEEE
  Transactions on}, vol.~21, no.~11, pp. 4662--4667, 2012.

\bibitem{Gupta}
M.~Gupta, S.~Narasimhan, and Y.~Schechner, ``On controlling light transport in
  poor visibility environments,'' in \emph{Computer Vision and Pattern
  Recognition, 2008. CVPR 2008. IEEE Conference on}.\hskip 1em plus 0.5em minus
  0.4em\relax IEEE, 2008, pp. 1--8.

\bibitem{Treibitz-Active}
T.~Treibitz and Y.~Schechner, ``Active polarization descattering,''
  \emph{Pattern Analysis and Machine Intelligence, IEEE Transactions on},
  vol.~31, no.~3, pp. 385--399, 2009.

\bibitem{Murez}
Z.~Murez, T.~Treibitz, R.~Ramamoorthi, and D.~Kriegman, ``Photometric stereo in
  a scattering medium,'' 2015.

\bibitem{Jagger}
W.~Jagger and W.~Muntz, ``Aquatic vision and the modulation transfer properties
  of unlighted and diffusely lighted natural waters,'' \emph{Vision research},
  vol.~33, no.~13, pp. 1755--1763, 1993.

\bibitem{Sch1}
Y.~Y. Schechner and N.~Karpel, ``Recovery of underwater visibility and
  structure by polarization analysis,'' \emph{Oceanic Engineering, IEEE Journal
  of}, vol.~30, no.~3, pp. 570 --587, july 2005.

\bibitem{Mort}
\BIBentryALTinterwordspacing
H.~Mortazavi, J.~Oakley, U.~of~Manchester. School~of Electrical, and
  E.~Engineering, \emph{Mitigation of Contrast Loss in Underwater
  Images}.\hskip 1em plus 0.5em minus 0.4em\relax University of Manchester,
  2010. [Online]. Available:
  \url{http://books.google.co.uk/books?id=L\_tsMwEACAAJ}
\BIBentrySTDinterwordspacing

\bibitem{Treibitz-Instant}
T.~Treibitz and Y.~Schechner, ``Instant 3descatter,'' in \emph{Computer Vision
  and Pattern Recognition, 2006 IEEE Conference on}, vol.~2, 2006, pp.
  1861--1868.

\bibitem{Mort3}
H.~Mortazavi and J.~Oakley, ``Underwater image enhancement by backscatter
  compensation,'' in \emph{Internation Conference on Modeling, Simulation and
  Applied Optimization. ICMSAO 2007.}\hskip 1em plus 0.5em minus 0.4em\relax
  IEEE, 2007.

\bibitem{Chandrasekhar}
S.~Chandrasekhar, \emph{Radiative transfer}, 1960.

\bibitem{Sun}
B.~Sun, R.~Ramamoorthi, S.~G. Narasimhan, and S.~K. Nayar, ``A practical
  analytic single scattering model for real time rendering,'' in \emph{ACM
  Transactions on Graphics (TOG)}, vol.~24, no.~3.\hskip 1em plus 0.5em minus
  0.4em\relax ACM, 2005, pp. 1040--1049.

\bibitem{Tsiotsios}
C.~Tsiotsios, M.~E. Angelopoulou, T.-K. Kim, and A.~J. Davison, ``Backscatter
  compensated photometric stereo with 3 sources,'' in \emph{Computer Vision and
  Pattern Recognition (CVPR), 2014 IEEE Conference on}.\hskip 1em plus 0.5em
  minus 0.4em\relax IEEE, 2014, pp. 2259--2266.

\bibitem{Kolagani}
N.~Kolagani, J.~S. Fox, and D.~R. Blidberg, ``Photometric stereo using point
  light sources,'' in \emph{Robotics and Automation, 1992. Proceedings., 1992
  IEEE International Conference on}.\hskip 1em plus 0.5em minus 0.4em\relax
  IEEE, 1992, pp. 1759--1764.

\bibitem{Tsiotsios-CVIU}
C.~Tsiotsios, T.~Kim, A.~Davison, and S.~Narasimhan, ``Model effectiveness
  prediction and system adaptation for photometric stereo in murky water,''
  \emph{Computer Vision and Image Understanding}, 2016.

\bibitem{Park}
J.~Park, S.~N. Sinha, Y.~Matsushita, Y.-W. Tai, and I.~S. Kweon, ``Calibrating
  a non-isotropic near point light source using a plane,'' in \emph{Computer
  Vision and Pattern Recognition (CVPR), 2014 IEEE Conference on}.\hskip 1em
  plus 0.5em minus 0.4em\relax IEEE, 2014, pp. 2267--2274.

\bibitem{OakleyBu}
J.~P. Oakley and H.~Bu, ``Correction of simple contrast loss in color images,''
  \emph{Image Processing, IEEE Transactions on}, vol.~16, no.~2, pp. 511 --522,
  feb. 2007.

\bibitem{Forsyth}
D.~A. Forsyth and J.~Ponce, ``A modern approach,'' \emph{Computer Vision: A
  Modern Approach}, 2003.

\bibitem{Frankot}
R.~T. Frankot and R.~Chellappa, ``A method for enforcing integrability in shape
  from shading algorithms,'' \emph{Pattern Analysis and Machine Intelligence,
  IEEE Transactions on}, vol.~10, no.~4, pp. 439--451, 1988.

\end{thebibliography}

%

\vspace{-13mm}

\begin{IEEEbiography}[{\includegraphics[width=1in,height=1.25in,clip,keepaspectratio]{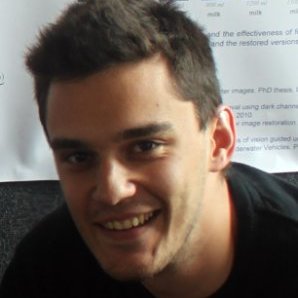}}]{Chourmouzios Tsiotsios}
has received the M.Eng. degree from the Department of Electrical and Computer Engineering, Aristotle University of Thessaloniki, Greece, and the PhD degree from the Department of Electrical and Electronic Engineering, Imperial College London, in 2010 and 2015, respectively. He is now a research fellow at Dyson Robotics Lab at Imperial College London working on 3D reconstruction and photometry.
\end{IEEEbiography}

\vspace{-15.5mm}

\begin{IEEEbiography}[{\includegraphics[width=1in,height=1.25in,clip,keepaspectratio]{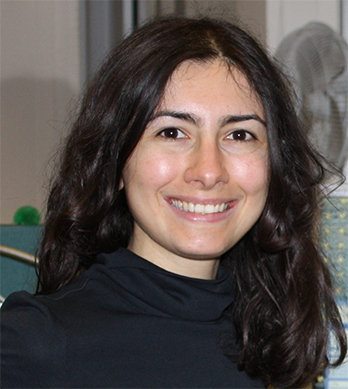}}]{Maria E. Angelopoulou}
	received the M.Eng. degree from the Department of Electrical and Computer Engineering, University of Patras, Greece, and the PhD degree from the Department of Electrical and Electronic Engineering, Imperial College London, in 2005 and 2008 respectively. In 2011, she was appointed as a researcher at CSP group at Imperial College London working on Photometric Stereo.
\end{IEEEbiography}

\vspace{-15.5mm}

\begin{IEEEbiography}[{\includegraphics[width=1in,height=1.25in,clip,keepaspectratio]{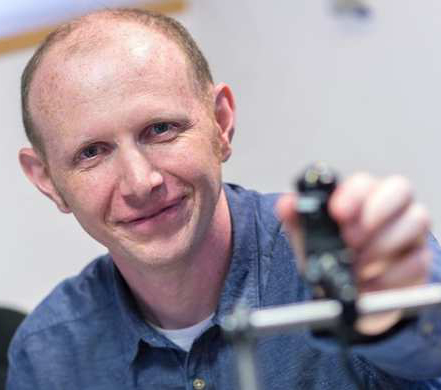}}]{Andrew J. Davison}
	is Professor of Robot Vision and Director of the Dyson
	Robotics Laboratory at Imperial College London. He has a B.A. physics and
	a D.Phil. in computer vision from the University of Oxford in 1994 and
	1998, respectively. In his doctorate at Oxford's Research Group he
	developed one of the first robot SLAM systems using vision. He joined Imperial College London as a Lecturer in 2005, held
	an ERC Starting Grant from 2008 to 2013 and was promoted to Professor
	in 2012. His Robot Vision Research Group continues to focus on
	advancing the basic technology of real-time localisation and mapping
	using vision, publishing advances in particular on real-time dense
	reconstruction and tracking, object-level mapping, and the use of novel sensing and
	processing in vision. He maintains a deep interest in exploring the
	limits of computational efficiency in real-time vision problems. In 2014 he became the founding Director of the new Dyson
	Robotics Laboratory at Imperial College, a lab working on the
	applications of computer vision to real-world domestic robots.
\end{IEEEbiography}

\vspace{-10mm}

\begin{IEEEbiography}[{\includegraphics[width=1in,height=1.25in,clip,keepaspectratio]{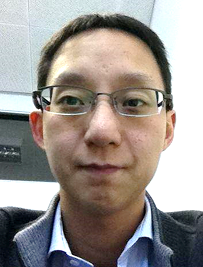}}]{Tae-Kyun Kim}
   is an Assistant Professor
   and leader of Computer Vision and Learning
   Lab at Imperial College London, UK, since Nov
   2010. He received the B.Sc. and M.Sc. degrees
   from Korea Advanced Institute of Science and
   Technology in 1998 and 2000, respectively, and
   worked at Samsung Advanced Institute of Technology
   in 2000-2004. He obtained his PhD from
   Univ. of Cambridge in 2008 and Junior Research
   Fellowship (governing body) of Sidney Sussex
   College, Univ. of Cambridge for 2007-2010. His
   research interests primarily lie in decision forests (tree-structure classifiers)
   and linear methods for: articulated hand pose estimation, face
   analysis and recognition by image sets and videos, 6D object pose
   estimation, active robot vision, activity recognition and object detection/tracking.
   He has co-authored over 40 academic papers in top-tier
   conferences and journals in the field, his co-authored algorithm for face
   image retrieval is an international standard of MPEG-7 ISO/IEC.
\end{IEEEbiography}





\end{document}